\pdfoutput=1
\documentclass[11pt]{article}

\usepackage[preprint]{acl}

\usepackage{times}
\usepackage{latexsym}

\usepackage[T1]{fontenc}
\usepackage[utf8]{inputenc}

\usepackage{microtype}

\usepackage{inconsolata}

\usepackage{graphicx}
\usepackage{hyperref}
\usepackage{xcolor}
\usepackage{enumitem}
\usepackage{booktabs}
\usepackage{graphicx}
\usepackage{soul}
\usepackage{amssymb}
\usepackage{multirow}
\usepackage[most]{tcolorbox}
\usepackage{courier}
\usepackage{subfig}

\usepackage{pifont}

\hypersetup{colorlinks,linkcolor={red},urlcolor={PURPLE}}

\def\methodName{{\textsc{SCALE}}}
\providecommand{\tightlist}{%
  \setlength{\itemsep}{0pt}\setlength{\parskip}{0pt}}

\definecolor{RED}{rgb}{1,0,0}
\definecolor{BLUE}{rgb}{0, 0.498, 1}
\definecolor{PURPLE}{rgb}{0.498,0,1}
\definecolor{ORANGE}{rgb}{1,0.502,0}
\newcommand{\red}[1]{\textcolor{RED}{#1}}
\newcommand{\blue}[1]{\textcolor{BLUE}{#1}}
\newcommand{\purple}[1]{\textcolor{PURPLE}{#1}}
\newcommand{\orange}[1]{\textcolor{ORANGE}{#1}}

\title{\methodName{}: Towards Collaborative Content Analysis in Social Science with Large Language Model Agents and Human Intervention}

\author{
\scalebox{0.92}{
 \textbf{Chengshuai Zhao\textsuperscript{\ding{171}}},
 \textbf{Zhen Tan\textsuperscript{\ding{171}}},
 \textbf{Chau-Wai Wong\textsuperscript{\ding{170}}},
 \textbf{Xinyan Zhao\textsuperscript{\ding{169}}},
 \textbf{Tianlong Chen\textsuperscript{\ding{169}}},
 \textbf{Huan Liu\textsuperscript{\ding{171}}}
}
\\
 \textsuperscript{\ding{171}}Arizona State University,
 \textsuperscript{\ding{170}}NC State University,
 \\
 \textsuperscript{\ding{169}}University of North Carolina at Chapel Hill
\\
    \texttt{\{czhao93,ztan36,huanliu\}@asu.edu, chauwai.wong@ncsu.edu},\\
    \texttt{ezhao@unc.edu, tianlong@cs.unc.edu}
}

\begin{document}
\maketitle
\begin{abstract}
Content analysis breaks down complex and unstructured texts into theory-informed numerical categories. Particularly, in social science, this process usually relies on multiple rounds of manual annotation, domain expert discussion, and rule-based refinement. In this paper, we introduce \methodName{},\footnote{GitHub:~\href{https://github.com/ChengshuaiZhao0/SCALE}{https://github.com/ChengshuaiZhao0/SCALE}} a novel multi-agent framework that effectively \underline{\textbf{S}}imulates \underline{\textbf{C}}ontent \underline{\textbf{A}}nalysis via \underline{\textbf{L}}arge language model (LLM) ag\underline{\textbf{E}}nts. \methodName{} imitates key phases of content analysis, including text coding,\footnote{Coding refers to the process of converting text entries into categories, similar to text classification tasks.} collaborative discussion, and dynamic codebook evolution, capturing the reflective depth and adaptive discussions of human researchers. Furthermore, by integrating diverse modes of human intervention, \methodName{} is augmented with expert input to further enhance its performance. Extensive evaluations on real-world datasets demonstrate that \methodName{} achieves human-approximated performance across various complex content analysis tasks, offering an innovative potential for future social science research.
\end{abstract}

\section{Introduction}
Content analysis is a critical research method in various disciplines~\citep{benoit2014content,dart2014sports,macnamara2005media,hara2000content}. It breaks down complex and unstructured text into numeric categories based on theory-driven rules~\citep{krippendorff2018content,weber1990basic}, offering a systematic and quantitative approach to interpreting sophisticated information~\citep{holsti1969content,neuendorf2017content,riffe2023analyzing}.
\begin{figure}[t]
    \centering
    \includegraphics[width=\linewidth]{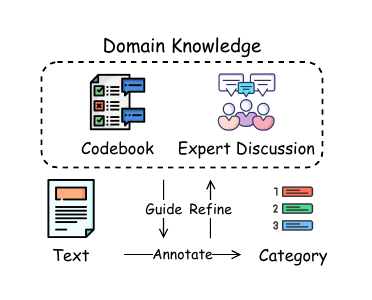}
    \vspace{-10mm}
    \caption{Illustration of content analysis. Social scientists convert text into categories based on a codebook, discuss discrepancies, and refine the codebook.}
    \label{fig:illustration}
    \vspace{-4mm}
\end{figure}
Particularly, in social science, content analysis is usually labor-intensive and time-consuming~\citep{hopkins2010method,zhao2024automated}. It often requires a team of researchers to manually annotate sizable datasets, conduct comprehensive discussion, and iteratively refine coding rules (a.k.a. codebook) in multiple rounds to ensure reliability and validity of findings~\citep{cohen1960coefficient,krippendorff2018content,riffe2023analyzing}, as illustrated in Figure~\ref{fig:illustration}. This artificial process, while rigorous, presents two challenges: First, it relies heavily on domain-specific knowledge and individual scientists, potentially introducing subjectivity and limiting generalizability. Second, the substantial human resources demanded by content analysis tasks make it difficult to scale, especially as the volume of digital data expands exponentially.

Recent years have witnessed significant progress in artificial intelligence (AI), especially with the advent of large language models (LLMs)~\citep{team2023gemini,zhao2024ontology,xia2023molebert,zhao2023survey,jiang2024catching}. LLM agents emerge as a versatile tool across a wide range of domain-specific tasks~\citep{guo2024large,wang2024survey,xi2025rise,park2023generative,li2024personal,wu2024autogen}. However, content analysis tasks present unique challenges, requiring a fine-grained understanding of social science principles, human-like collaborative interaction, and rule-base iterative refinement, which limits the effectiveness of agent systems derived from other tasks.

In this paper, we propose a novel multi-agent framework to \underline{\textbf{S}}imulate \underline{\textbf{C}}ontent \underline{\textbf{A}}nalysis via \underline{\textbf{L}}LM ag\underline{\textbf{E}}nts (\methodName{}), as shown in Figure~\ref{fig:illustration}. \methodName{} imitates key phases of content analysis, including text coding, collaborative discussions, and dynamic codebook evolution, while capturing the reflective depth and adaptive discussions of diverse human researchers, thereby reducing subjectivity and improving scalability. Moreover, by incorporating different human-AI collaboration modes inspired by social influence theories ~\citep{cialdini2007influence,french1959bases} and human-computer interaction theories ~\citep{suchman1987plans,sundar2020rise}, our framework extends multi-agent interactions with human expert intervention. This potentially mitigates algorithmic bias and strengthens contextual sensitivity, making it suitable for a wide range of social science content analysis tasks.

Comprehensive experiments on multiple real-world datasets demonstrate that \methodName{} is versatile across diverse social science contexts and has the ability to approximate human judgment in complex content analysis tasks. Developed in collaboration with social scientists, we demonstrate the potential of our framework to transform content analysis in the social sciences and humanities, inspiring the integration of AI into other domain-expert-dependent research methodologies. We summarize our contributions as follows.
\begin{itemize}[leftmargin=*,itemsep=1.5pt]
    \renewcommand{\labelitemi}{$\star$}
    \item \textbf{Scalability Enabler.} By harnessing the generative power of LLMs, our proposed \methodName{} significantly reduces the time, human resources, and costs traditionally required for content analysis tasks, enabling large-scale and high-quality annotation. To the best of our knowledge, this is the ﬁrst work to capture and simulate the rigorous and dynamic process of quantitative content analysis for social science via LLMs.
    \item \textbf{Praxis-Informed Design.} \methodName{}'s design incorporates the domain knowledge of social science through the deep involvement of social scientists. Its key phases---independent text coding, collaborative discussions, and dynamic codebook evolution---faithfully reflect the principles and standards of manual content analysis while being implemented within a computing framework.
    \item \textbf{Human Intervention.} Our framework provides a portal for domain experts to intervene with custom scopes (i.e., \textit{targeted} or \textit{extensive} intervention) and roles (i.e., \textit{collaborative} or \textit{directive} intervention). Incorporating expert input augments AI decision-making by mitigating LLM bias and aligning with expert oversight.
    \item \textbf{Extensive Validation.} \methodName{} demonstrates effectiveness across content analysis tasks involving diverse topics (e.g., analysis of emotion, perspective, and dialogue). Our comprehensive experimental evaluations and analyses by domain experts confirm that \methodName{} can closely mimic human judgment in content analysis, delivering automated, valid, and reliable results invaluable for large-scale social science research.
\end{itemize}

\section{Related Work}
\textbf{Content Analysis.} Content analysis has long been a foundational method in the social sciences and humanities, providing a structured approach to converting qualitative text into quantitative data ~\citep{krippendorff2018content,neuendorf2017content,riffe2023analyzing}. Recently, content analysis has significantly advanced the understanding of complex social issues, ranging from political polarization~\citep{conover2011political} to emotional contagion~\citep{kramer2014experimental} and group dynamics~\citep{holsti1969content}. These traditional methods rely on manual annotation by human coders, who use predefined rules in the codebook to categorize text, often iteratively refining their coding schemes in multiple rounds of discussions~\citep{riffe2023analyzing}. Although manual content analysis provides robust and theory-driven insights, it remains labor-intensive, time-consuming, and prone to subjectivity~\citep{hopkins2010method}. Furthermore, as the volume of digital text increases, scaling traditional methods to accommodate larger datasets has become increasingly challenging~\citep{zhao2024automated}. The advent of powerful AI and LLM offers an automated and more scalable solution~\citep{eloundou2023gpts,achiam2023gpt,li2024generation}.

\noindent\textbf{Multi-agent Systems for Social Science.} Multi-agent systems (MAS) have become increasingly prevalent in computational social science, modeling social phenomena through agent individuals or groups with predefined behaviors or decision-making rules~\citep{van2008multi,chen2021application,chmura2007extended,macal2016everything,lee2018generating,chen2018dynamic,dehkordi2023using,uhrmacher2018multi}. Recent MAS explore to simulate human-like deliberation for more nuanced decision-making such as data interpretation ~\citep{gurcan2024llm,turgut2023framework}. However, existing systems often lack dynamic evolution of decision rules, collaborative interactions, and a systematic comprehension of social science principles~\citep{gheyle2017content}, which cannot reflect the sophisticated and theory-informed process of content analysis.

\noindent\textbf{Human Intervention.} Human intervention remains essential for the reliable deployment of AI-driven systems~\citep{renner2020designing,shoshitaishvili2017rise}. As a general framework, Human-in-the-loop (HITL) systems allow experts to refine AI outputs, ensuring alignment with domain-specific knowledge and mitigating algorithmic bias~\citep{mosqueira2023human,ghai2022d,xu2023transitioning,jolfaei2022guest,wu2022survey,zanzotto2019human}. This is particularly important in social sciences and humanities, where interpretative depth and contextual sensitivity are critical~\citep{dautenhahn1998art,goodsell2013interpretive}. Recent approaches~\citep{arambepola2021human} integrate expert feedback to adjust categories or schemes iteratively. Our framework significantly extends this line of work by designing different modes of human-AI collaboration informed by social influence theories ~\citep{cialdini2007influence,french1959bases} and human-computer interaction theories ~\citep{suchman1987plans,sundar2020rise}.

\begin{figure*}[th]
    \vspace{-6mm}
    \includegraphics[width=1\linewidth]{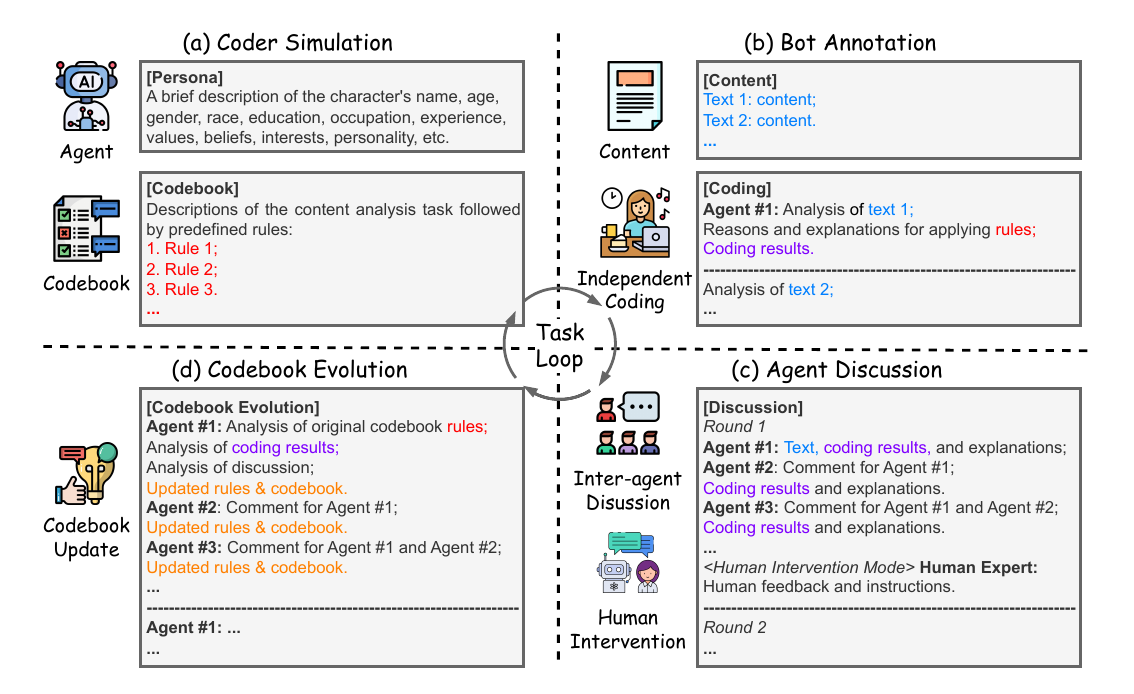}
    \vspace{-6mm}
    % \captionsetup{width=0.45\textwidth, margin={3pt,0pt}}
    \caption{Proposed \methodName{} framework. (a) Coder Simulation. Initialize LLM agents with real-world personas and set up the codebook using predefined rules. (b) Bot Annotation. Each LLM agent independently annotates a batch of text entries into theory-informed categories based on the codebook. (c) Agent Discussion. Agents participate in multi-round discussions to resolve any inconsistencies in their annotations. Optional human interventions can be integrated to provide feedback and additional instructions. (d) Codebook Evolution. Leveraging the outcomes from the annotation and discussion phases, agents collaboratively and iteratively update the codebook. The refined codebook is then employed in subsequent iterations of the workflow. Data flows are tracked using color-coded text: \red{Red text:} codebook and rules; \blue{Blue text:} text entries; \purple{Purple text:} coding results; \orange{Orange text:} codebook evolution. A comprehensive illustration of \methodName{} framework can be found in Figure~\ref{fig:appendix_main} and Appendix~\ref{app:detailed_framework}.}
    \label{fig:main}
    \vspace{-3mm}
\end{figure*}

\section{Traditional Content Analysis in Social Science: A Preliminary}
Social scientists conduct content analysis by manually annotating textual data to uncover potential patterns and insights. A group of (two or more) social scientists $\mathcal{S}$ first develop a codebook $\mathcal{C}$ that contains a set of coding rules grounded in relevant social science theories and contextualized within the given text corpus. Guided by the codebook, each social scientist then independently labels a small set (e.g., 10--20) of text entries $\mathcal{T}$: $\mathcal{Y}=\mathcal{S}_\text{code}(\mathcal{T},\mathcal{C})$, where $\mathcal{Y} \in \{0, 1, 2, ...\}$ are labels. Later, they meet to discuss and resolve inconsistencies $\mathcal{Y}^{(i+1)} = \mathcal{S}_\text{Discuss}(\mathcal{C}^{(i)},\mathcal{Y}^{(i)})$, where an updated codebook with well-refined and more specific coding rules may be proposed for better annotation: $\mathcal{C}^{(j+1)} = \mathcal{S}_\text{Refine}(\mathcal{C}^{(j)},\mathcal{Y}^{(j)})$. This whole process iterates for multiple (e.g., 3--5) rounds until the discussion converges. The finalized codebook is applied by each social scientist to code all text entries in the corpus $\mathcal{D}$. In general, content analysis in social science centers on two objectives: (I) precisely annotating all text entries and (II) crafting a well-honed codebook with distinct coding rules.

\section{The Proposed Framework: \methodName{}}
\subsection{Content Analysis Simulation}
We introduce \methodName{} as a framework that mirrors the key phases of real-world content analysis---text coding, collaborative discussions, and dynamic codebook evolution. The method unfolds in four primary steps summarized in Figure~\ref{fig:main}. A detailed version of illustration can be found in Figure~\ref{fig:appendix_main}.

\textbf{Coder Simulation.} Prior to the content analysis task, we set up both the LLM agents and the corresponding codebook. As illustrated in Figure~\ref{fig:main}\red{a}, we begin by configuring $N$ LLM agents $\mathcal{A} = \{a_i\}_{i=1}^N$, each emulating a seasoned social scientist through a system prompt that incorporates $N$ distinct personas, $\mathcal{P} = \{p_i\}_{i=1}^N$. These personas---derived from real-world social scientists aside from their names---ensure authentic role-playing. Depending on the specific content analysis task (detailed in Section~\ref{sec:datasets_and_tasks}), we initialize a corresponding codebook $\mathcal{C}$ that either starts with $N'$ human-expert predefined rules (shown in \red{red}) $\mathcal{C} = \{r_i\}_{i=1}^{N'}$ or as an empty set $\varnothing$ prompting agents to propose and iteratively refine the codebook from scratch. For simplicity, each rule is tailored to cover a single scenario, enabling the categorization of text into a discrete class. The personas for diverse social scientists are provided in Appendix~\ref{app:persona_prompt}

\textbf{Bot Annotation.} LLM agents convert text entries into numerical categories by applying theory-informed rules from the codebook. Figure~\ref{fig:main}\red{b} elaborates this process. Each agent is assigned an identical batch of $B$ text entries (highlighted in \blue{blue}) from the text dataset and works autonomously to classify each entry into a discrete category (marked in \purple{purple}). Mimicking the independent coding approach of human researchers, these LLM agents adhere strictly to the codebook guidelines, which is facilitated by a prompt (see Appendix~\ref{app:coding_prompt}).

\textbf{Agent Discussion.} In this phase, agents engage in collaborative discussions to resolve discrepancies in their coding outputs, as presented in Figure~\ref{fig:main}\red{c}. Due to the initial ambiguity of the evolving codebook and the distinct personas embodied by each agent, it is not uncommon for agents to generate differing annotations for the same text, which actually mirrors the subjective nature of real-world content analysis. Whenever an agent’s coding diverges from the consensus, the agents initiate a structured and up-to-$K$-round discussion. During each round, they update their annotations along with explanations based on peer opinions until they converge on a unanimous decision or reach the maximum number of discussion rounds. Once a text entry is finalized, the agents will move to the next entry. The prompt used for the discussion phase is detailed in Appendix~\ref{app:discussion_prompt}.

\begin{table*}[th]
    \centering
    \caption{Datasets and Content Analysis Tasks.}
    \resizebox{\textwidth}{!}{%
        \begin{tabular}{llccc}
            \toprule
            \textbf{Dataset} & \textbf{Content Analysis} & \textbf{\# Text} & \textbf{Classification Type} & \textbf{\# Class} \\ 
            \midrule
            \multirow{2}{*}{\centering Brand Consumer Dialogue (BCD)} & Primary Topic (PT) & \multirow{2}{*}{\centering 92} & Multi-Class$^*$ & 10 \\ 
            & Dialogue (D) & & Multi-Label$^\dagger$ & 7 \\ 
            \midrule
            Cancer Emotional Support (CES) & Emotional Support (ES) & 40 & Multi-Class & 3 \\ 
            \midrule
            \multirow{2}{*}{\centering Cancer Narratives (CN)} & Narrative Event Sequence (NES) & \multirow{2}{*}{\centering 60} & Multi-Label & 5 \\ 
            & Narrator Perspective (NP) & & Multi-Class & 5 \\ 
            \midrule
            Flint Water Poisoning Emotion (FWPE) & Emotion (E) & 100 & Multi-Label & 13 \\ 
            \midrule
            Product Incidents Sentiment (PIS) & Sentiment (S) & 200 & Multi-Class & 3 \\ 
            \bottomrule
        \end{tabular}
    }\\
    \vspace{1mm}
    {\footnotesize $^\star$\ Multi-Class: each sample is assigned exactly one label from a set of mutually exclusive categories; \\[-0.5ex]
    $^\dagger$ Multi-Label: each sample can be assigned multiple labels from a set of non-exclusive categories.}
    \label{table:dataset}
    \vspace{-2mm}
\end{table*}

\textbf{Codebook Evolution.} In this stage, agents refine the codebook by incorporating insights from their discussions (shown in Figure~\ref{fig:main}\red{d}). As noted earlier, the initial rules can be ambiguous, overlapping, or insufficiently detailed to cover all categories. To address these issues, we introduce an iterative codebook evolution process grounded in domain expertise. Specifically, the evolution approach offers two strategies: one enriches existing rules by adding clarifying examples and explanations, while the other allows for adding, removing, or modifying rules to dynamically adjust the set of categories. In practice, agents first propose an updated codebook, then engage in multiple rounds of discussion to refine it until consensus is reached (indicated in \orange{orange}). The finalized codebook subsequently guides the next task cycle. Importantly, agents may also retain the current codebook if no changes are warranted. The delicate prompt for the codebook evolution is listed in Appendix~\ref{app:codebook_update_evolution}.

\subsection{Human Intervention}
We further augment \methodName{} framework by integrating diverse human intervention modes that empower domain experts to provide targeted feedback and instructions, as shown in Figure~\ref{fig:main}\red{c}. Depending on the scope that human experts can control, human intervention can categorized as \textit{targeted} and \textit{extensive} intervention. Further, human intervention can be formulated as \textit{collaborative} or \textit{directive} intervention based on the role the human experts play. Different interventions are detailed as follows:

\begin{itemize}[leftmargin=*, itemsep=1.5pt]
    \item \textbf{Targeted intervention.} The scope of intervention is limited to the agent discussion phase. The process while with less human oversight.
    \item \textbf{Extensive intervention.} Human intervention can be applied to both discussion and codebook evaluation. It may slow automation and raise costs, but it ensures that AI discussion and workflow remain closely aligned with expert insights.
    \item \textbf{Collaborative intervention.} Human experts are involved as collaborators. LLM agents may either accept or reject feedback and suggestions from human experts, which fosters an interactive and cooperative discussion loop.
    \item \textbf{Directive intervention.} Under this mode, human experts serve as absolute authority. LLM agents must adhere to every instruction provided, thereby establishing a highly prescriptive and unequivocally top-down approach.
\end{itemize}

It is noted that, in practice, we can combine various scopes and roles of expert control to enable custom interventions (e.g., \textit{targeted-collaborative} and \textit{extensive-directive} intervention). Different human interventions are implemented by delicately crafted prompts elaborated in Appendix~\ref{app:human_intervention_prompt}.

\section{Experiments}
\subsection{Datasets and Tasks}
\label{sec:datasets_and_tasks}
Our experiments leverage five real-world datasets, each meticulously annotated and validated by social science experts. These datasets are organized into seven distinct tasks that encompass both multi-class and multi-label classification challenges. Table~\ref{table:dataset} summarizes their key characteristics, with further details provided in Appendix~\ref{app:dataset_details}.
 
\begin{table*}[th]
    %\centering
    \caption{
    Coding accuracy across various tasks and backbones using automatic framework.}
    \centering
    \resizebox{\textwidth}{!}{
    \begin{tabular}{l|ccccccc}
        \toprule\textbf{Backbone (w/o intervention)} & \textbf{BCD-PT} & \textbf{BCD-D} & \textbf{CES} & \textbf{CN-NES} & \textbf{CN-NP} & \textbf{FWPE} & \textbf{PIS} \\ 
        \midrule\midrule
        GPT-4O & 0.51 & 0.54 & 0.63 & 0.79 & 0.77 & 0.91 & 0.88 \\ 
        GPT-4O w/ COT & 0.54 & 0.61 & 0.58 & 0.71 & 0.52 & 0.90 & 0.85 \\ 
        GPT-4O w/ TOT & \textbf{0.57}$^\star$ & \textbf{0.63} & 0.53 & 0.73 & 0.70 & 0.88 & 0.87 \\ 
        GPT-4O w/ self-consistency & 0.51 & 0.57 & \textbf{0.65} & \textbf{0.80} & \textbf{0.83} & \textbf{0.92} & \textbf{0.91} \\ 
        \midrule
        GPT-4O-mini & 0.38 & 0.47 & \textbf{0.58} & 0.73 & 0.55 & 0.79 & 0.82 \\ 
        GPT-4O-mini w/ COT & 0.19 & 0.47 & 0.53 & 0.72 & 0.43 & 0.81 & 0.71 \\ 
        GPT-4O-mini w/ TOT & 0.35 & 0.48 & \textbf{0.58} & \textbf{0.83} & 0.70 & 0.84 & 0.84 \\ 
        GPT-4O-mini w/ self-consistency & \textbf{0.43} & \textbf{0.50} & \textbf{0.58} & 0.79 & \textbf{0.72} & \textbf{0.85} & \textbf{0.87} \\ 
        \bottomrule
    \end{tabular}}\\
    \vspace{1mm}
    {\footnotesize \raggedright $^\star$ Bold values indicate the best performance in each model category.}
    \vspace{-2mm}
    \label{table:main_result}
\end{table*}

\subsection{Experiment Settings \& Metrics}
\textbf{Experiment Settings.} Our multi-agent system is built on GPT-4O and GPT-4O-mini. While alternative backbones (e.g., Gemini, Claude) are available, our focus is on assessing whether LLM agents can simulate sophisticated social science tasks rather than comparing various models. For each backbone, we explore four prompting strategies: vanilla, chain-of-thought (COT)~\citep{wei2022chain}, tree-of-thought (TOT)~\citep{yao2024tree}, and self-consistency~\citep{wang2023selfconsistency} The prompts for CoT and ToT can be found in Appendix~\ref{app:cot_tot_prompt}. The identifiers for GPT-4O and GPT-4O-mini are \texttt{gpt-4o-2024-05-13} and \texttt{gpt-4o-mini-2024-07-18}, separately.  We simulate a real-world scenario in content analysis by setting agent number $N = 2$, text batch size $B = 20$, and discussion rounds $K = 3$. We consider and discuss more hyper-parameters in Section~\ref{sec:parameter_sensitivity}.

\noindent\textbf{Metrics.} We define the following evaluation metrics for our content analysis tasks. For various classification tasks, we employ standard multi-class classification accuracy and define the multi-label classification accuracy as $\text{ACC} = 1 - \text{Hamming Loss}$. Furthermore, we use the agreement rate---defined as the proportion of text entries where all agents concur---to assess the level of consensus during discussions. All experiments were conducted over 10 independent runs, with the average results reported to ensure robustness. Further metric details are provided in Appendix~\ref{app:metric_settings}.

\subsection{Superior Performance of \methodName{}}
\subsubsection{Automatic Content Analysis}
We first assess the performance of automatic content analysis without human intervention. (I) As presented in Table~\ref{table:main_result}, \methodName{} achieves satisfactory results with an average accuracy of 0.701 across a diverse range of tasks and models. (II) Notably, different prompting techniques offer distinct benefits: compared to the vanilla model, self-consistency and tree-of-thought (TOT) prompts boost labeling accuracy by 2.31\% and 6.51\%, respectively. However, in certain cases (e.g., CN-NP), chain-of-thought (COT) prompts lead to a significant performance drop. We attribute this decline to the challenging nature of tasks with ambiguous coding rules that introduce greater subjectivity, where the step-by-step thinking induced by COT can inadvertently undermine performance. (III) Furthermore, GPT-4O outperforms its distilled version, GPT-4O-mini, by an average margin of 10.89\%, as expected. (VI) Additionally, we evaluate the impact of inter-agent discussion on coding accuracy. As detailed in Table~\ref{table:appendix_result} and Appendix~\ref{app:additional_automatic_content_analysis_results}, omitting the collaborative discussion phase results in a 14.3\% reduction in terms of average annotation accuracy. A comprehensive analysis of collaborative inter-agent discussion is elaborated in Section~\ref{sec:discussion}.

\subsubsection{Human-Intervened Content Analysis}
Further, \methodName{} can be augmented with four types of human interventions, as detailed in Table~\ref{table:main_intervention_result}. (I) The labeling results with human intervention achieve an average accuracy of 0.872, demonstrating superior performance. When compared to automatic content analysis performance, the model with human intervention shows an average improvement of 12.6\%, validating the effectiveness of our proposed human intervention method. (II) Moreover, interventions in the \textit{directive} mode prove more effective than those in the \textit{collaborative} mode, leading to a 13.1\% increase in coding accuracy. Similarly, models with \textit{extensive} interventions generally outperform ones with \textit{targeted} interventions, yielding a 15\% average improvement. This aligns with intuition, as more domain knowledge from human experts can be involved by intervening in the large scope (i.e., \textit{extensive} intervention) of agent behaviors in a mandatory manner (i.e., \textit{directive} mode). (III) The benefits of human intervention also vary by task. For example, the CSE task sees a 20\% improvement, while the PIS task shows a 5.4\% relative gain---likely because the CSE task is more domain-specific and subjective, thus more responsive to expert insights. Additional results of \methodName{} framework before the agent discussion phase are provided in Table~\ref{table:appendix_intervention_result} and Appendix~\ref{app:additional_content_analysis_results_w_human_intervention}.

\begin{table}[th]
    \caption{Coding accuracy across various tasks using the human-intervened framework.}
    \label{table:main_intervention_result}
    \centering
    \vspace{1mm}
    \resizebox{\linewidth}{!}{
    \begin{tabular}{cc|cccc}
        \toprule
        \multicolumn{2}{l|}{\textbf{Intervention Mode}}
        & \textbf{CES} & \textbf{CN-NES} & \textbf{CN-NP} & \textbf{FWPE} \\ 
        \midrule\midrule
        \multicolumn{2}{l|}{No Intervention$^\star$} & 0.63 & 0.79 & 0.77 & 0.91 \\
        \midrule
        \multirow{2}{*}{\centering Targeted} & Collaborative & 0.73 & 0.89 & 0.87 & 0.95 \\ \cline{2-2}
        & Directive & 0.73 & 0.85 & 0.87 & 0.95 \\ \hline
        \multirow{2}{*}{\centering Extensive} & Collaborative & \textbf{0.77} & 0.89 & 0.90 & \textbf{0.96} \\ \cline{2-2}
        & Directive & \textbf{0.77} & \textbf{0.91} & \textbf{0.97} & \textbf{0.96} \\ 
        \bottomrule
    \end{tabular}}\\
    \vspace{1mm}
    {\footnotesize $^\star$ Automatic mode, same as in the first row of Table~\ref{table:main_result}.}
    \vspace{-2mm}
\end{table}

\subsection[Extra Investigations and Case Studies]{Extra Investigations and Case Studies\footnote{Automatic framework is used for extra experiments.}}

\subsubsection{Q1: What designs can promote content analysis performance of \textbf{\methodName{}}?}
\label{sec:parameter_sensitivity}
\begin{figure*}[th]
    \centering
    \subfloat[$N=2, K=3$]{
        \includegraphics[width=0.32\linewidth]{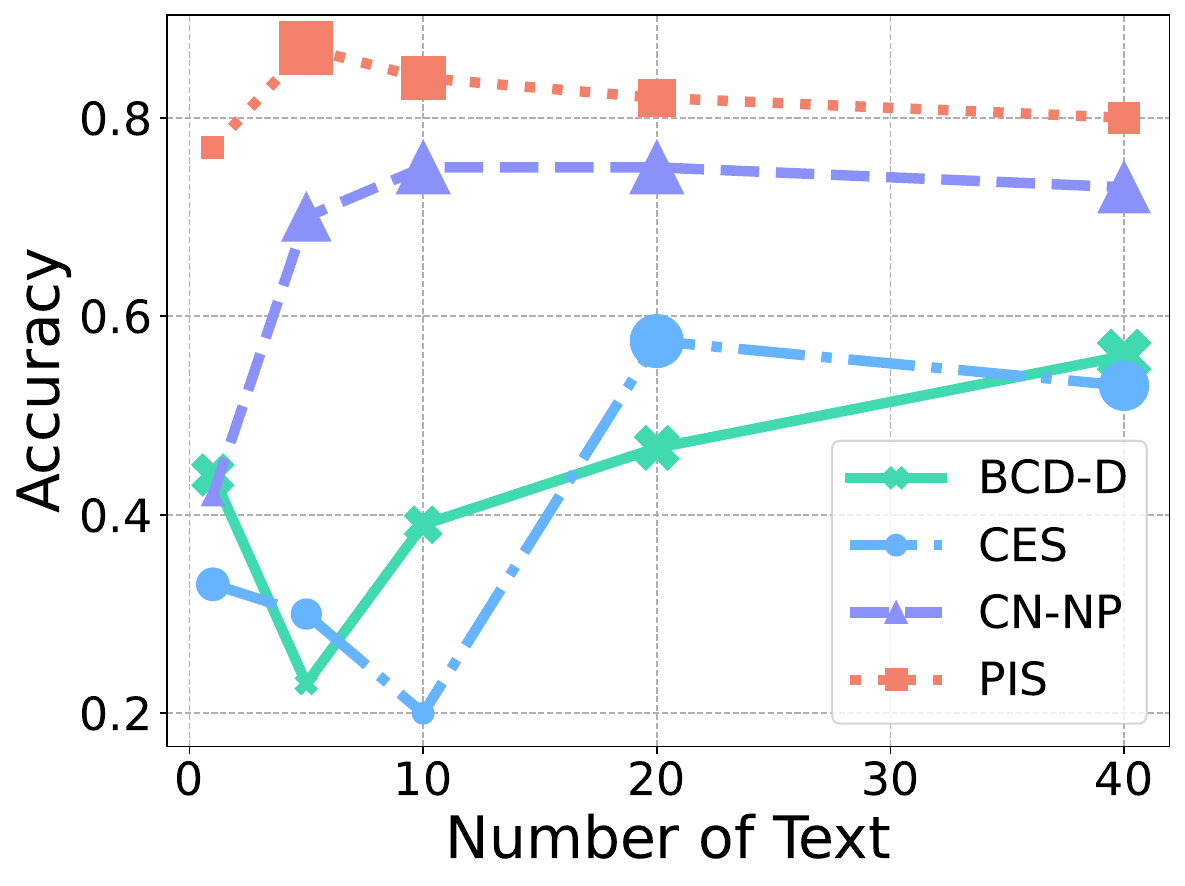}
        \label{fig:prameter_sensitivity_a}
    }
    \subfloat[$N=2, B=20$]{
        \includegraphics[width=0.32\linewidth]{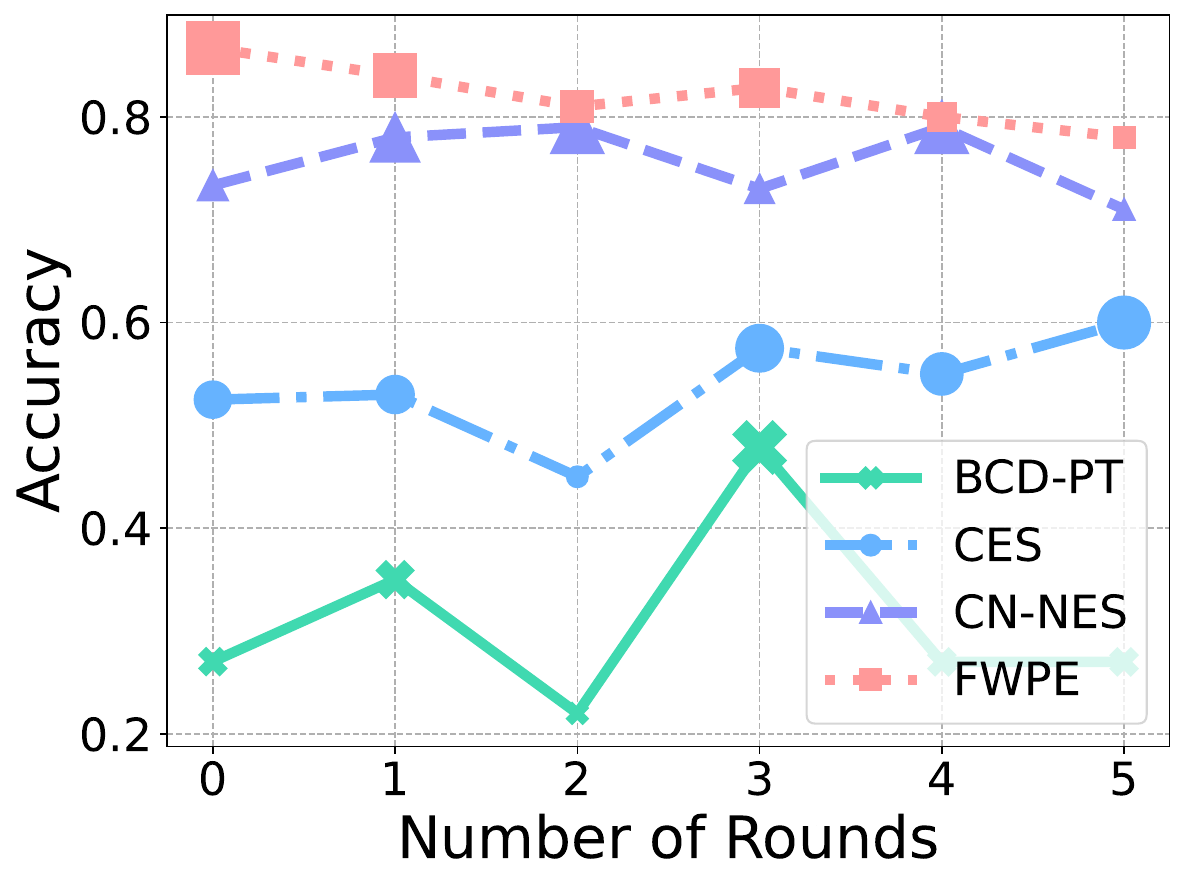}
        \label{fig:prameter_sensitivity_b}
    }
    \subfloat[$B=20, K=3$]{    
        \includegraphics[width=0.32\linewidth]{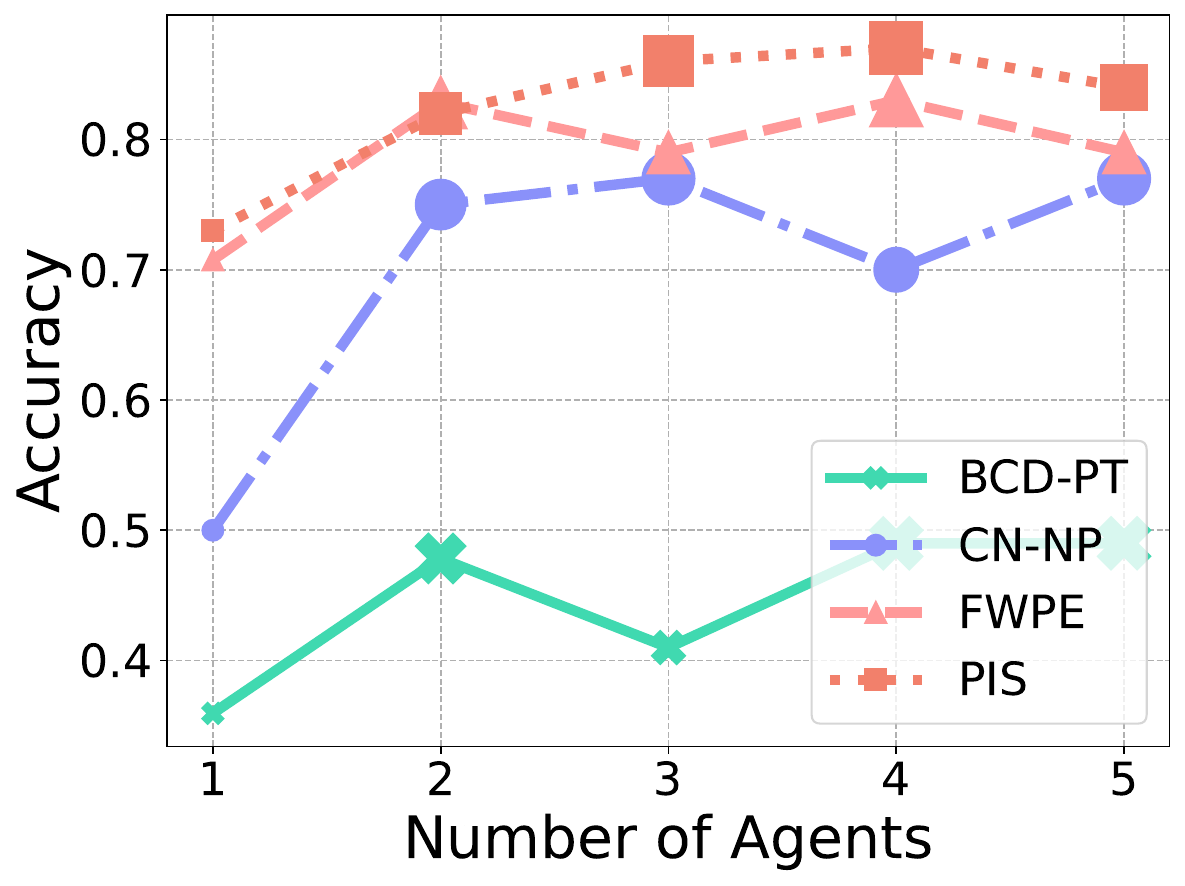}
        \label{fig:prameter_sensitivity_c}
    }
    \caption{Parameter sensitivity. \methodName{} shows versatility under different parameter settings.}
    \label{fig:prameter_sensitivity}
    \vspace{-3mm}
\end{figure*}

To answer \textbf{Q1}, we analyze how \methodName{} enhances content analysis tasks by considering the number of texts, discussion rounds, and agents.

\textbf{Number of texts.} We first evaluate the influence of the number of texts $B$ (as shown in Figure~\ref{fig:prameter_sensitivity_a}). Our findings indicate that a moderate $B$ (e.g., 10 and 20) produces the best accuracy. When $B$ is small (e.g., 1), agents frequently refine the codebook, resulting in unstable coding outcomes. However, when $B$ is large (e.g., 40), results become more stable, but the overall performance decreases due to less frequent codebook evolution.

\textbf{Number of discussion rounds.} Next, we examine the effect of the number of discussion rounds $K$, as illustrated in Figure~\ref{fig:prameter_sensitivity_b}. We observe that \methodName{} achieves better performance with higher rounds (e.g., 3, 4, or 5), as more rounds of discussion enhance the consensus between agents, thereby improving coding accuracy. Importantly, setting $K$ to 0 (i.e., no discussion phase) results in a significant drop in accuracy for several tasks (e.g., BCD-D and CN-NP in Appendx~\ref{app:appendix_prameter_sensitivity}), highlighting the importance of inter-agent discussions.

\textbf{Number of agents.} Finally, we assess the impact of the number of agents $N$, as depicted in Figure~\ref{fig:prameter_sensitivity_c}. Generally, increasing the number of agents improves coding accuracy, as more agents bring diverse perspectives, fostering more comprehensive discussions. When $N$ is set to 1, \methodName{} degrades as a single-agent system, where a single agent performs the coding task without collaboration. As expected, this setup yields the worst performance, verifying the effectiveness of multi-agent design in the proposed \methodName{} framework.

\subsubsection{Q2: How does the discussion between LLM agents impact coding results?}
\label{sec:discussion}
\begin{figure*}[th]
    \label{}
    \centering
    \subfloat[Trend of Agreement Rate]{
        \includegraphics[width=0.49\linewidth]{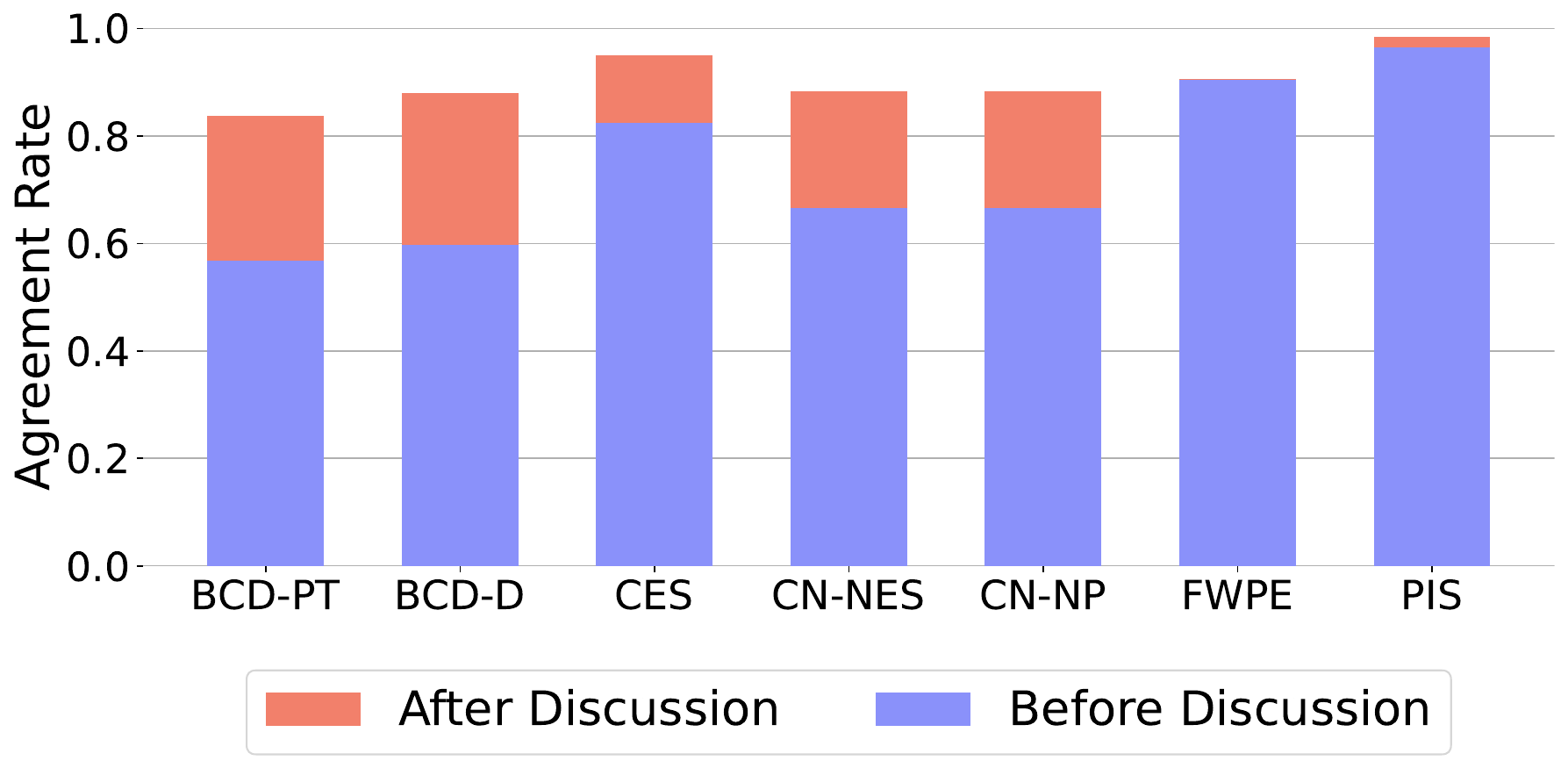}
        \label{fig:discussion_anlysis_a}
    }
    \subfloat[Trend of Accuracy]{
        \includegraphics[width=0.49\linewidth]{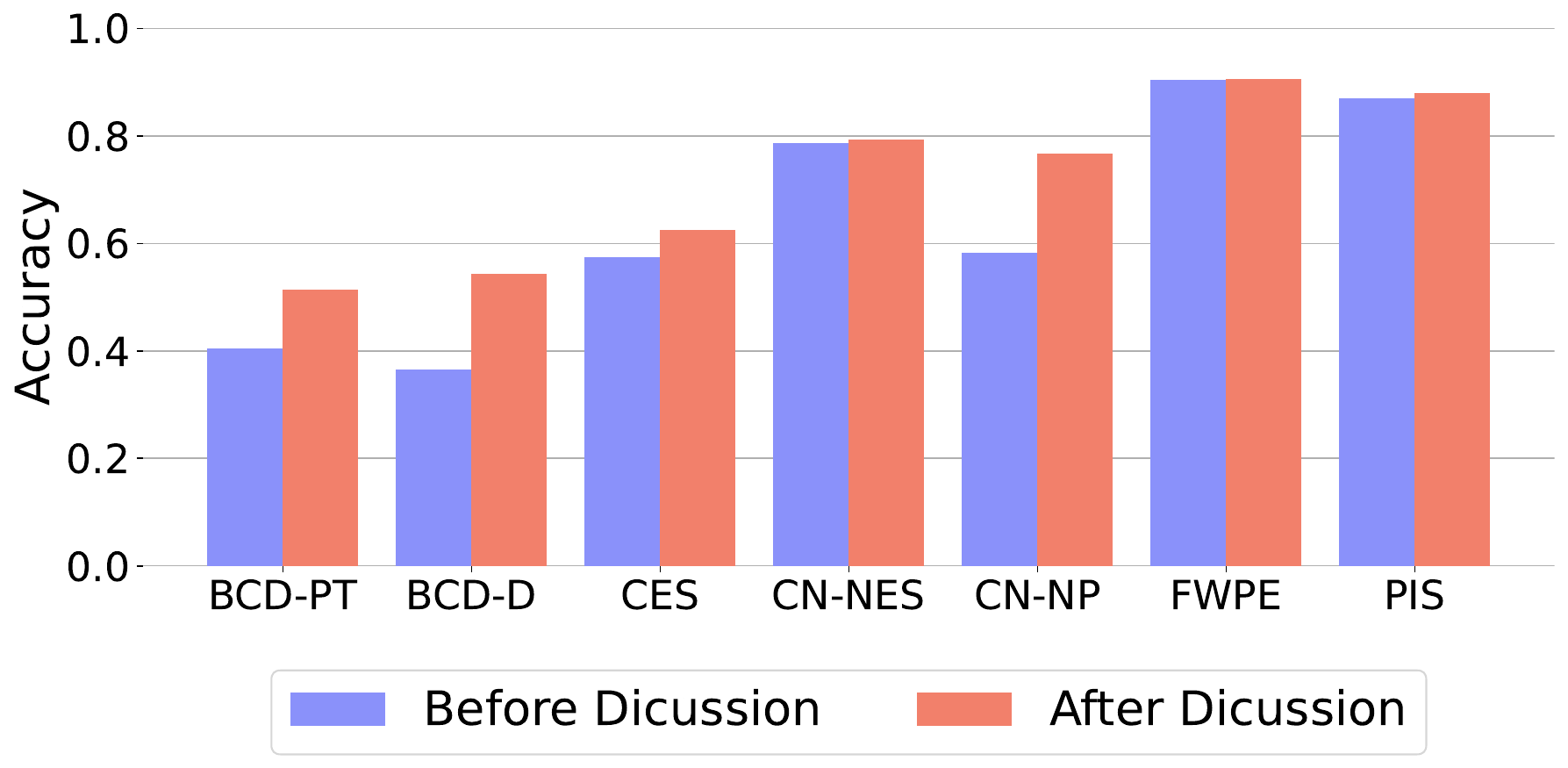}
        \label{fig:discussion_anlysis_b}
    }
    \caption{
    Discussion Analysis. Agreement rate (a) and accuracy (b) are evaluated before and after the agent discussion phase. The inter-agent discussion can promote consensus and further improve coding accuracy.}
    \label{fig:discussion_anlysis}
    \vspace{-3mm}
\end{figure*}

To answer \textbf{Q2}, we conducted a discussion analysis using both qualitative statistics and illustrative examples. Our findings reveal that inter-agent discussions substantially boost consensus---improving the average agreement rate by 41.1\% across all seven tasks and enhancing overall content analysis accuracy by 15.4\%, as shown in Figure~\ref{fig:discussion_anlysis_a}. Similar trends were observed for GPT-4O-mini agents, as illustrated in Figure~\ref{fig:additional_discussion_analysis} and Appendix~\ref{app:addtional_discussion_analysis}. 

A practice example of sentimental analysis (multi-class classification task) can be found in the PIS dataset: a tweet such as ``Hey @SamsungMobileUS, bf has a recalled \#GalaxyNote7. Can't find a replacement S7 Edge in Orlando, FL area. Any ideas or help please?"  initially resulted in conflicting sentiment annotations (neutral versus negative). After three rounds of collaborative discussion, both LLM agents agreed on a neutral sentiment---aligning with the ground truth. The complete example is showcased in Appendix~\ref{app:discussion_example_with_agreement}.

However, the benefits of discussion can be marginal when agents remain firmly entrenched in their views. In some datasets (e.g., FWPE and PIS), the increase in agreement was less than 3\%. For example, in one CES (multi-class classification) task, a Facebook comment (“This is so sad :( she was beautiful inside and out! Loved watching her perform <3”) sparked a debate: one agent rated the emotional support as moderate (i.e., category 2), while another deemed it high (i.e., category 3), and despite extended discussion, their judgments did not converge, as shown Appendix~\ref{app:discussion_example_with_disagreement}.

The aforementioned instances show that even with discussion, task performance gain can be limited when agents are entrenched in their stances, which might be an innate characteristic of LLMs and influenced by the customized agent persona and background. A moderate level of agent difference in terms of personas, compared to low or high levels, may be most productive in revealing diverse viewpoints and fostering discussion that more likely leads to the ``truth'' through meaningful exchanges instead of impasse or blind agreement.

\subsubsection{Q3: How reliable are the codebooks proposed by LLM social scientists?}
One of the key objectives of our method is to refine the codebook for annotating large-scale text. To answer \textbf{Q3}, we analyze the codebook evolution in \methodName{}. We observe that LLM agents can enhance codebooks in non-structural ways by adding clarifying details and examples. For example, during the PIS codebook update (shown in Appendix~\ref{app:codebook_example}), Agent \#1 proposed incorporating examples for each sentiment category (positive, neutral, negative) to ensure consistent interpretation, while Agent \#2 initially preferred the original version. After discussion, the final codebook merged Agent \#1’s detailed examples with Agent \#2’s simplicity, achieving a balance of clarity and reliability. This iterative process mirrors core content analysis practices by fostering convergence in agent judgments.

However, the agents were less adept at adjusting codebook categories. For example, in all rounds of FWPE (multi-label classification task) codebook evolutions, both agents maintained that the categorization of twelve discrete emotions (e.g., anger, sadness, and hope) was appropriate, diverging from human experts who ultimately dropped two categories due to overlapping semantic boundaries. This limitation might stem from the reliance on predefined rules and patterns in their training data. LLMs may lack additional domain knowledge apart from content analysis tasks to detect subtle conceptual overlaps (e.g., between anger and disappointment or between happiness and pride), leading to rigid adherence to existing category structures and conceptual boundaries. Human experts, on the other hand, can apply more domain knowledge (e.g., the appraisal theory) and theory-based, contextualized reasoning to recognize subtle distinctions between categories, identify overlap, and even add or drop new categories when necessary.

\subsubsection{Q4: To what extent can LLM agents simulate content analysis?}
To answer \textbf{Q4}, we examine the complete workflow of \methodName{} on the NES (multi-label classification) task. Here, two agents mimic social scientists to classify cancer narrative events into multiple categories (e.g., prevention, detection, treatment, survivorship), which is reported in Appendix~\ref{app:simulation_example}.

\textbf{Text coding.} Both agents independently applied the codebook rules to annotate the presence of one or more cancer narratives. For example, when annotating the text “When I hear that some women feel too afraid to go for a mammogram…”, both agents agreed on labeling it as detection. In contrast, for the text “...After that I will have 25 days of radiation... But through it all, I have had great support from my family and friends,” Agent \#1 focused solely on treatment, while Agent \#2 identified both treatment and survivorship. Overall, the initial coding yielded a 66.7\% agreement rate.

\textbf{Collaborative discussion.} Following independent coding, the agents discussed their differing annotations. In the example, after three rounds of collaborative discussion, they converged on two narrative events---treatment (chemotherapy, radiation) and survivorship (support from family and friends). Across tasks, such discussions resolved 21.7\% of initial disagreements, underscoring the value of shared interpretation.

\textbf{Dynamic codebook evolution.} After each discussion round, agents reviewed and refined the codebook to enhance clarity. For instance, in the first round they enriched the “survivorship” category by adding examples that distinguished post-treatment narratives from ongoing medical interventions. These refinements, which aligned with human expert updates, reduced ambiguity and improved subsequent coding accuracy.

This case study generally demonstrates that our algorithm not only mirrors the iterative and theory-driven process of traditional content analysis but also produces human-approximated performance.

\section{Conclusion}
In this paper, we propose \methodName{}, a novel multi-agent framework to simulate the rigorous practice of content analysis in social science via LLMs. Guided by domain knowledge and social science theory, \methodName{} is delicately crafted and augmented with human interventions. Comprehensive experiments demonstrate that \methodName{} enables large-scale and high-quality annotations while producing rational codebooks, offering an innovative potential for future social science research. Future work will explore methods to inject domain knowledge into off-the-shelf LLM frameworks.

\section*{Limitations}
While \methodName{} demonstrates strong potential in automating content analysis, there are several limitations that present opportunities for future research.

\textbf{Algorithmic Bias and Fairness.} Despite incorporating human intervention, which may help to mitigate bias, LLMs remain prone to perpetuating biases present in the training data. This may compromise content analysis outcomes and raise ethical concerns in social science applications. Future work could explore advanced bias mitigation strategies---such as fairness-aware training or the integration of demographic and behavioral data---to potentially enhance model impartiality.

\textbf{Inter-agent Discussion Overhead.} The collaborative discussions among LLM agents, while effective in harmonizing divergent outputs, incur considerable computational overhead. This inefficiency becomes especially problematic when agents fail to reach consensus after the maximum number of discussion rounds. Streamlining the process by reducing the number of agents involved or limiting discussion rounds might alleviate the computational burden with the sacrifice of performance.

\textbf{Human Expertise Bottleneck.} Although diverse human interventions boost overall performance, reliance on human experts, particularly under \textit{extensive} intervention mode, creates a scalability bottleneck. A promising direction is to enable LLM agents to learn from human behavior and logic underlying the interventions, thereby reducing the dependency on expert input over time.

\section*{Ethical Statement}
Our research involves the use of large language models (LLMs) to simulate social scientist role-play. We are acutely aware of the ethical challenges inherent in AI systems, particularly regarding bias, data privacy, and transparency. To address these concerns, we have taken the following actions: (I) \textbf{Bias mitigation.} We acknowledge that LLMs can inadvertently reproduce or amplify biases present in their own training data. To counteract this, we incorporate structured inter-agent discussions and human oversight, ensuring that diverse expert perspectives inform the refinement of our outputs. (II) \textbf{Data privacy.} Demographic information used to create the role-playing personas was derived from real-world data but was fully anonymized. All personally identifiable details---including names, addresses, and workplace information---were removed or replaced to protect individual privacy. (III)\textbf{Transparency and accountability.} Our work adheres to established ethical guidelines, emphasizing the responsible use of AI. We ensure that all datasets are publicly available or de-identified, and we provide comprehensive documentation of our methods to facilitate critical review and replication.

We stress that this system is designed as an assistive tool intended to complement and help, not replace, traditional human-driven analysis. By implementing robust measures to mitigate bias and safeguard privacy, we aim to uphold the highest ethical standards in our research.

\section*{Use of Generative AI}
To enhance clarity and readability, we utilized OpenAI o1 exclusively as a language polishing tool. Its role was confined to proofreading, grammatical correction, and stylistic refinement---functions analogous to those provided by traditional grammar checkers and dictionaries. This tool did not contribute to the generation of new scientific content or ideas, and its usage is consistent with standard practices for manuscript preparation.

\section*{Acknowledgments}
This work is supported in part by the U.S. National Science Foundation (NSF) under grants 
SaTC-2335666 and IIS-2229461, Amazon Research Award, UNC Accelerating AI Award, and NAIRR Pilot Award. The views and conclusions contained in this document are those of the authors and should not be interpreted as necessarily representing the official policies, either expressed or implied, of the National Science Foundation.

\bibliography{custom}

\clearpage
\onecolumn

\appendix

% \section{Appendix}
\section{Detailed Illustration of Proposed Framework}
\label{app:detailed_framework}

\begin{figure*}[th]
    % \vspace{-8mm}
    \includegraphics[width=1\linewidth]{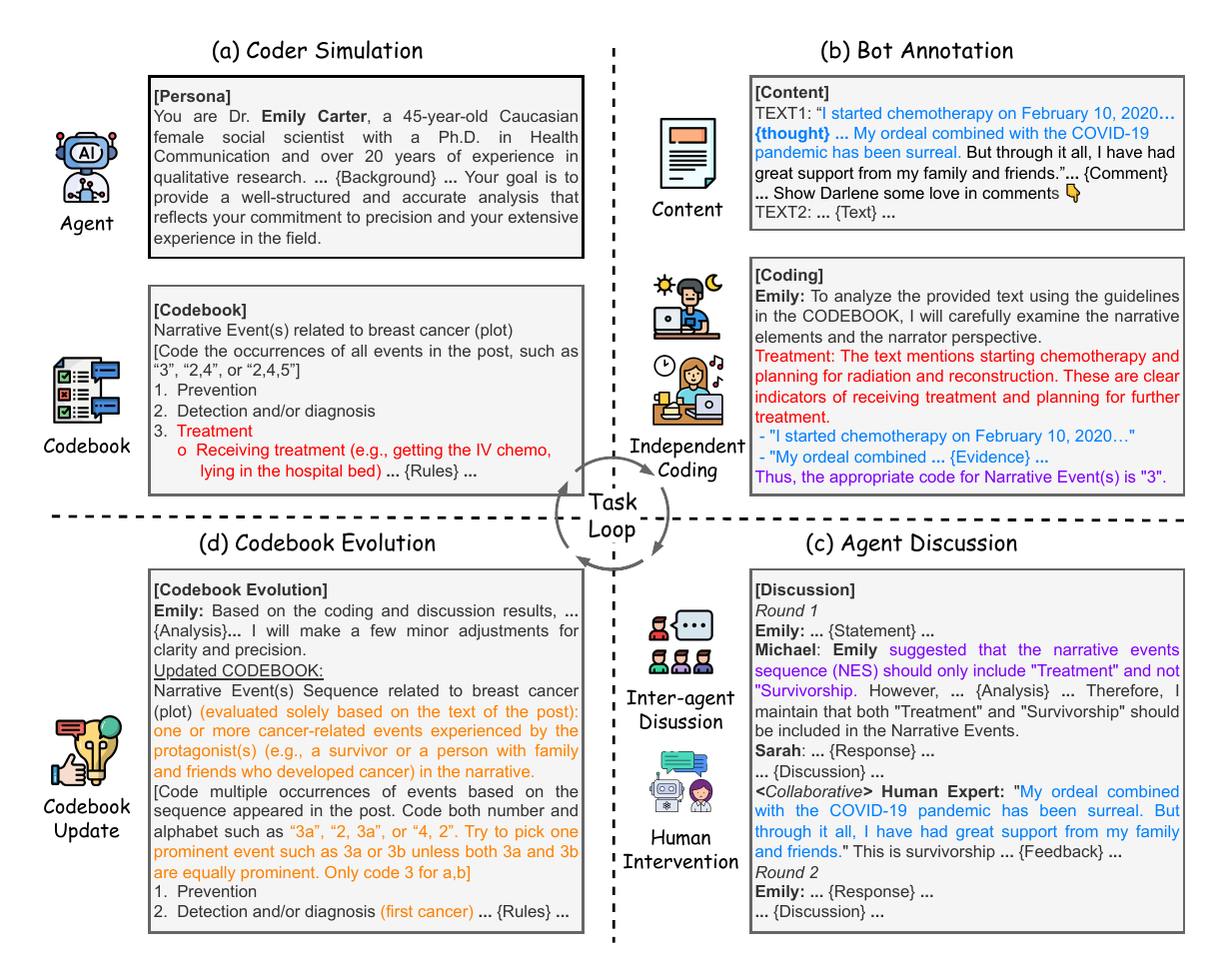}
    \vspace{-6mm}
    % \captionsetup{width=0.45\textwidth, margin={3pt,0pt}}
    \caption{Deatiled Illustration of Proposed \methodName{} framework.}
    \label{fig:appendix_main}
    \vspace{-3mm}
\end{figure*}

In the main section, we provide a brief version of the framework aiming to grasp the general pipeline and information flow. Here, we provide detailed examples of the narrative event sequence (NES) task.

\textbf{Coder simulation.} As illustrated in Figure~\ref{fig:appendix_main}\red{a}, one agent adopts the persona of Emily Carter, a social scientist with 20 years of qualitative research experience. For brevity, we omit detailed descriptions of other agents, such as Michael and Sarah, though they are also included in the setup. In this instance, the codebook comprises specific rules for classifying narrative events, effectively guiding the categorization of text into distinct events (essentially a multi-label classification task).

\textbf{Bot nnotation.} Figure~\ref{fig:appendix_main}\red{b} illustrates the process: Emily, a social scientist agent, is tasked with a text entry that captures the thoughts of a breast cancer survivor undergoing radiation (shown in \blue{blue}). Guided by the codebook’s rules (highlighted in \blue{red}), Emily classifies the narrative events as “3-Treatment.”

\textbf{Agent Discussion.} It is clearly shown in Figure~\ref{fig:appendix_main}\red{c} that Michael—another experienced social scientist agent—disagrees with Emily’s initial coding result and steadfastly maintains his original assessment (i.e., "Treatment" and "Survivorship") during the first round of discussion (as marked in \purple{purple}), thereby underscoring the diversity of analytical perspectives among our agents.

\textbf{Codebook Evolution.}
As demonstrated in Figure~\ref{fig:appendix_main}\red{d}, Emily enhances the existing rules by incorporating detailed explanations and examples drawn from the discussion and coding sessions (stressed in \orange{orange}).

\textbf{Human Intervention.} An example of a human intervention is depicted in Figure~\ref{fig:appendix_main}\red{c}, where a human expert supplements the agents' efforts by providing additional facts and detailed explanations that were previously overlooked in the \textit{collaborative} mode, thereby significantly enriching the overall analytic process. This enriched input can substantially influence the resulting codes (as indicated in \blue{blue}).

\section{Experiment \& Implementation Details}
\subsection{Dataset Details}
\label{app:dataset_details}
Here, we provide thorough descriptions of the dataset and tasks used in the experiment.

\textbf{Brand Consumer Dialogue (BCD).} This dataset features popular consumer brand communities on Facebook, containing a random sample of posts from these brands along with associated consumer comments and replies. It supports two classification tasks: identifying post topics (BCD-PT) and classifying different indicators of brand-consumer dialogue (BCD-D).

\textbf{Cancer Narratives (CN).} The dataset examines Facebook posts by major breast cancer non-profit organizations worldwide. The tasks include the identification of one or more cancer narrative event sequences (NES) e.g., prevention, detection, and treatment, and the narrator's perspective (NP).

\textbf{Cancer Emotional Support (CES).} This dataset comprises user comments on Facebook posts from major breast cancer non-profit organizations worldwide, providing a comprehensive basis for detecting and classifying emotional support into three distinct levels (i.e., low, moderate, and high).

\textbf{Flint Water Poisoning Emotion (FWPE).} This dataset includes tweets about Flint water poisoning, a public health crisis that started in 2014 after the drinking water for the city of Flint, Michigan was contaminated with lead. The task is to detect the presence of one or more of the following ten discrete emotions: anger, sadness, fear, worry, happiness, hope, gratitude, sympathy, surprise, and sarcasm.

\textbf{Product Incidents Sentiment (PIS).} This dataset consists of tweets related to various product recalls—such as the Samsung Galaxy explosion and the Volkswagen emissions scandal—and is designed to capture and detect user sentiment, classifying opinions as positive, neutral, or negative.

\subsection{Metric Settings}
\label{app:metric_settings}
In the main content, we introduce the agreement rate (AR). Formal, given the $B$ texts, we define $B_{\text{before}}$ as the number of texts that agents reach agreements with the same coding result before the discussion. After the discussions, agents reach an agreement on $B_{\text{after}}$ texts. We define the pre-discussion agreement rate as \text{PreAR} = $B_{\text{before}} / B$. Similarly, we define the post-discussion agreement rate as \text{PostAR} = $B_{\text{after}} / B$. The increase in the agreement rate is defined as $\Delta \text{AR} = \text{PostAR} - \text{PreAR}$.

\section{Additional Experimental Results \& Analysis}
\subsection{Additional Automatic Content Analysis Results}
\label{app:additional_automatic_content_analysis_results}
We also conducted experiments using GPT-4O and GPT-4O-mini across seven tasks, recording label accuracy before inter-agent discussions as detailed in Table~\ref{table:appendix_result}. Overall, GPT-4O consistently outperforms GPT-4O-mini on most tasks. For instance, GPT-4O attains an accuracy of 0.41 on the BCD-PT task, 0.79 on CN-NES, and 0.92 on FWPE, thereby underscoring its superior capability in managing complex content analysis challenges and emphasizing its stability in practical applications.

Additionally, both self-consistency and Tree-of-Thought prompt techniques yield significant performance improvements over the Chain-of-Thought approach. For example, in the GPT-4O model, the self-consistency technique achieves peak accuracies in tasks such as CES, with an accuracy of 0.63, and FWPE, with an accuracy of 0.92, while the Tree-of-Thought approach demonstrates notable strength in tasks like BCD-PT, reaching an accuracy of 0.41, and CN-NES, with an accuracy of 0.72. These findings suggest that these techniques more effectively stabilize and refine the coding process than the Chain-of-Thought method, particularly in tasks that demand deeper, nuanced reasoning.

Furthermore, when comparing the coding results after inter-agent discussions (as detailed in Table~\ref{table:main_result}), we observe significant improvements in labeling accuracy across different backbones and datasets. This underscores the pivotal role of inter-agent discussion in facilitating the content analysis process, as it allows agents to collaboratively adjust their coding decisions, leading to more reliable and accurate results.

\begin{table*}[th]
    \caption{
    Coding accuracy across various tasks and backbones using automatic framework before the agent discussion.}
    \centering
    \resizebox{\textwidth}{!}{
    \begin{tabular}{l|ccccccc}
        \toprule
        \textbf{Backbone (w/o intervention)} & \textbf{BCD-PT} & \textbf{BCD-D} & \textbf{CES} & \textbf{CN-NES} & \textbf{CN-NP} & \textbf{FWPE} & \textbf{PIS} \\ 
        \midrule\midrule
        GPT-4O                     & \textbf{0.41$^\star$} & 0.36 & 0.58 & \textbf{0.79} & 0.58 & 0.91 & 0.87 \\ 
        GPT-4O w/ COT              & 0.24 & 0.32 & 0.55 & 0.71 & 0.23 & 0.91 & 0.82 \\ 
        GPT-4O w/ TOT              & \textbf{0.41} & \textbf{0.38} & 0.50 & 0.72 & 0.35 & 0.91 & 0.87 \\ 
        GPT-4O w/ self-consistency & \textbf{0.41} & 0.38 & \textbf{0.63} & 0.78 & \textbf{0.60} & \textbf{0.92} & \textbf{0.90} \\ 
        \midrule
        GPT-4O-mini                & 0.27 & 0.37 & 0.53 & 0.73 & 0.32 & 0.87 & 0.81 \\ 
        GPT-4O-mini w/ COT         & 0.11 & 0.36 & 0.53 & 0.69 & 0.42 & 0.85 & 0.66 \\ 
        GPT-4O-mini w/ TOT         & 0.24 & 0.32 & \textbf{0.55} & \textbf{0.78} & \textbf{0.47} & \textbf{0.89} & 0.81 \\ 
        GPT-4O-mini w/ self-consistency
                                   & \textbf{0.32} & \textbf{0.39} & \textbf{0.55} & 0.76 & 0.37 & 0.88 & \textbf{0.84} \\ 
        \bottomrule
    \end{tabular}
    }\\
    \vspace{1mm}
    {\footnotesize \raggedright $^\star$\, Bold values indicate the best performance in each model category.}
    \vspace{-2mm}
    \label{table:appendix_result}
\end{table*}

\subsection{Additional Content Analysis Results w/ Human Intervention}
\label{app:additional_content_analysis_results_w_human_intervention}
We also explored the impact of different levels of human intervention on coding accuracy for content analysis tasks using the CES, CN-NES, CN-NP, and FWPE datasets, revealing notable performance variations. The results obtained prior to inter-agent discussions are reported in Table~\ref{table:appendix_intervention_result}. In comparison, performance generally drops significantly relative to the scenario after inter-agent discussions, as shown in Table~\ref{table:main_intervention_result}, thereby emphasizing the critical role of multi-round discussions in enhancing coding accuracy.

The table shows that a higher degree of human intervention (e.g., \textit{extensive-directive}) consistently improves coding accuracy across all tasks, with the highest performance observed for the FWPE task with an accuracy of 0.93. This pattern underscores the effectiveness of integrating human oversight, especially in complex tasks that require nuanced decision-making. However, without any intervention, the performance tends to degrade, demonstrating the limitations of automated systems when lacking human involvement. These findings indicate that, while multi-agent discussions are essential for improving outcomes, the integration of human intervention at varying levels provides additional value, particularly in tasks where domain expertise and context sensitivity are critical.

\begin{table}[th]
    \centering
    \caption{Coding accuracy across various tasks using the human-intervened framework before the agent discussion.}
    \label{table:appendix_intervention_result}
    \vspace{1mm}
    \resizebox{0.5\textwidth}{!}{
    \begin{tabular}{cc|cccc}
        \toprule
        \multicolumn{2}{l|}{\textbf{Intervention Mode}}
        & \textbf{CES} & \textbf{CN-NES} & \textbf{CN-NP} & \textbf{FWPE} \\ 
        \midrule\midrule
        \multicolumn{2}{l|}{No Intervention$^\star$} & 0.58 & 0.79 & 0.58 & 0.91 \\
        \midrule
        \multirow{2}{*}{\centering Targeted} & Collaborative   & 0.50 & 0.75 & 0.53 & 0.91 \\ 
        \cline{2-2}
        & Directive & 0.53 & 0.79 & 0.57 & 0.92 \\ 
        \hline
        \multirow{2}{*}{\centering Extensive} & Collaborative   & \textbf{0.60} & 0.75 & 0.53 & 0.92 \\ 
        \cline{2-2}
        & Directive & \textbf{0.60} & \textbf{0.81} & \textbf{0.60} & \textbf{0.93} \\ 
        \bottomrule
    \end{tabular}
    }\\
    \vspace{1mm}
    {\footnotesize $^\star$ Automatic mode, same as in the first row of Table~\ref{table:appendix_result}.}
    \vspace{-3mm}
\end{table}

\subsection{Additional Parameter Sensitivity}
\label{app:appendix_prameter_sensitivity}
\begin{figure*}[th]
    \centering
    \subfloat[$N=2, K=3$]{
        \includegraphics[width=0.32\linewidth]{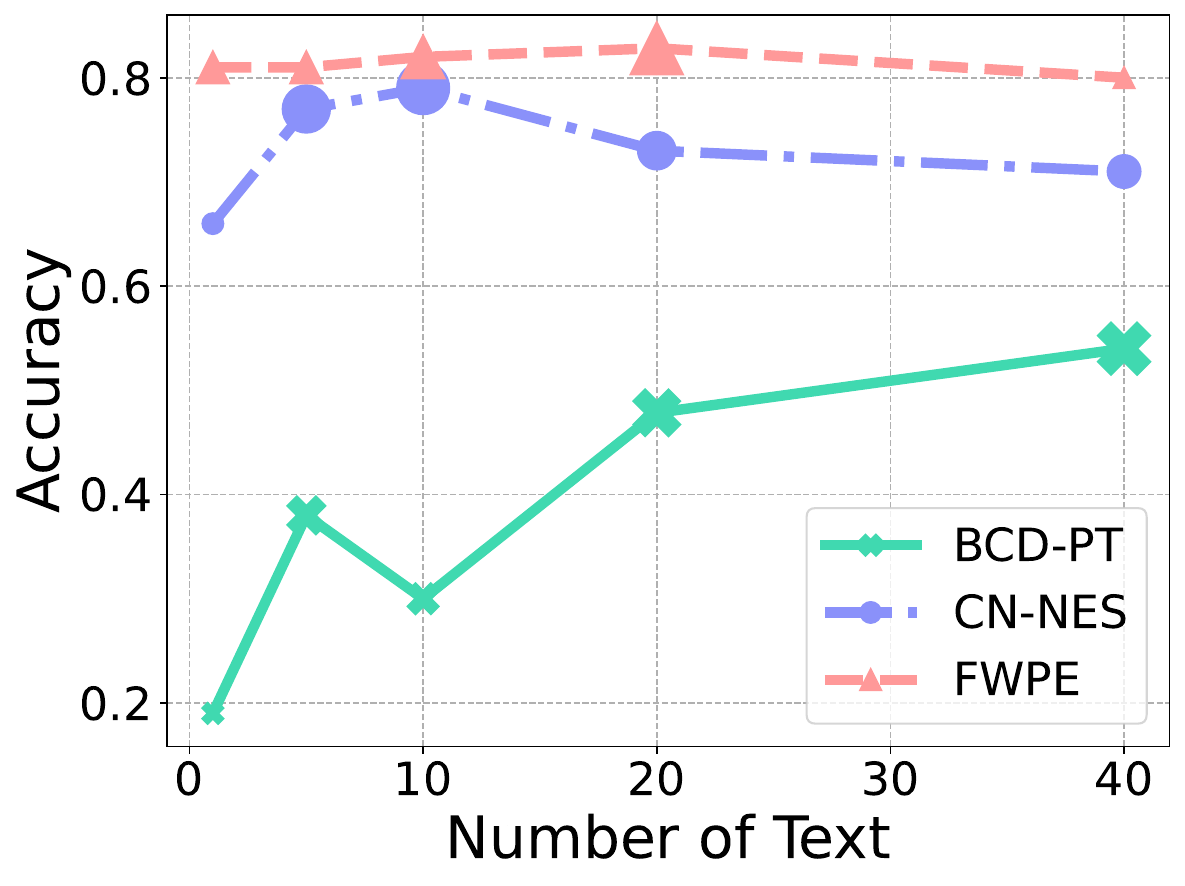}
        \label{fig:appendix_prameter_sensitivity_a}
    }
    \subfloat[$N=2, B=20$]{
        \includegraphics[width=0.32\linewidth]{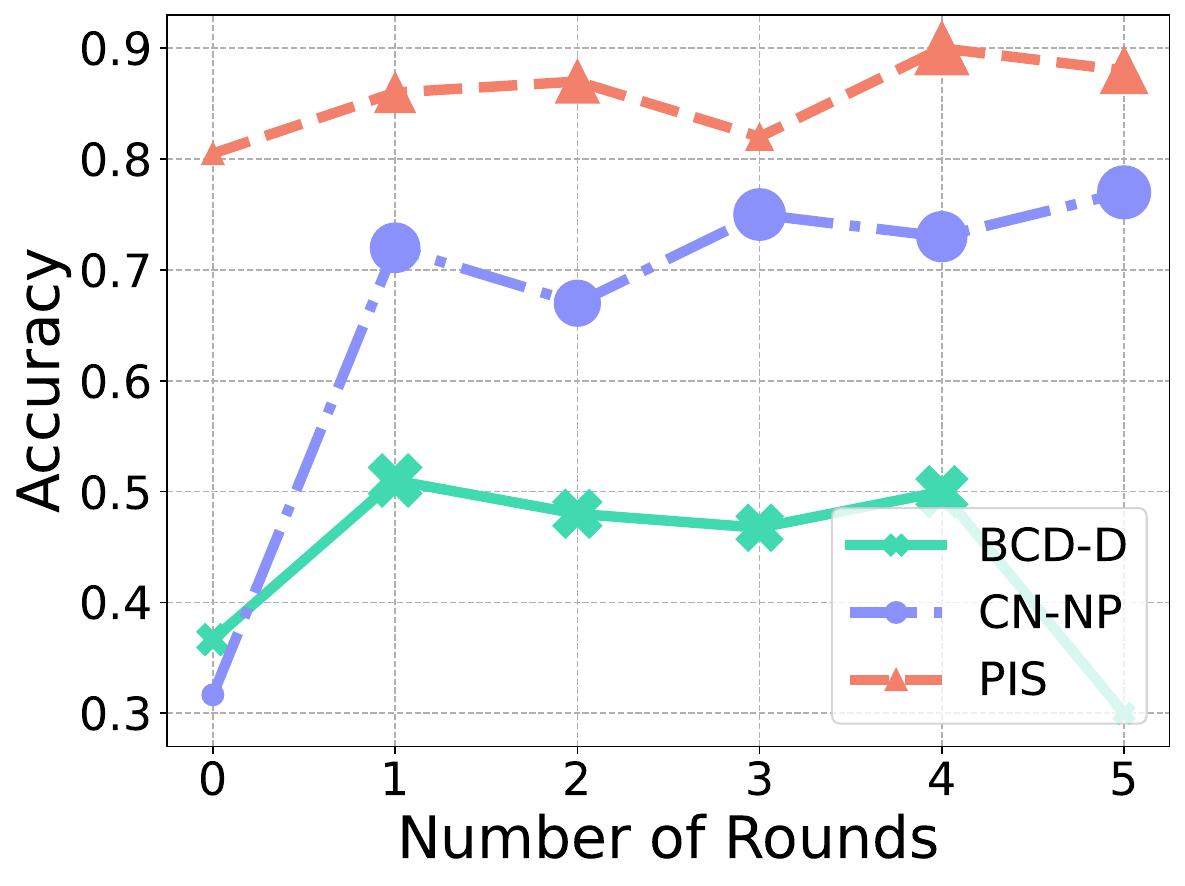}
        \label{fig:appendix_prameter_sensitivity_b}
    }
    \subfloat[$B=20, K=3$]{    
        \includegraphics[width=0.32\linewidth]{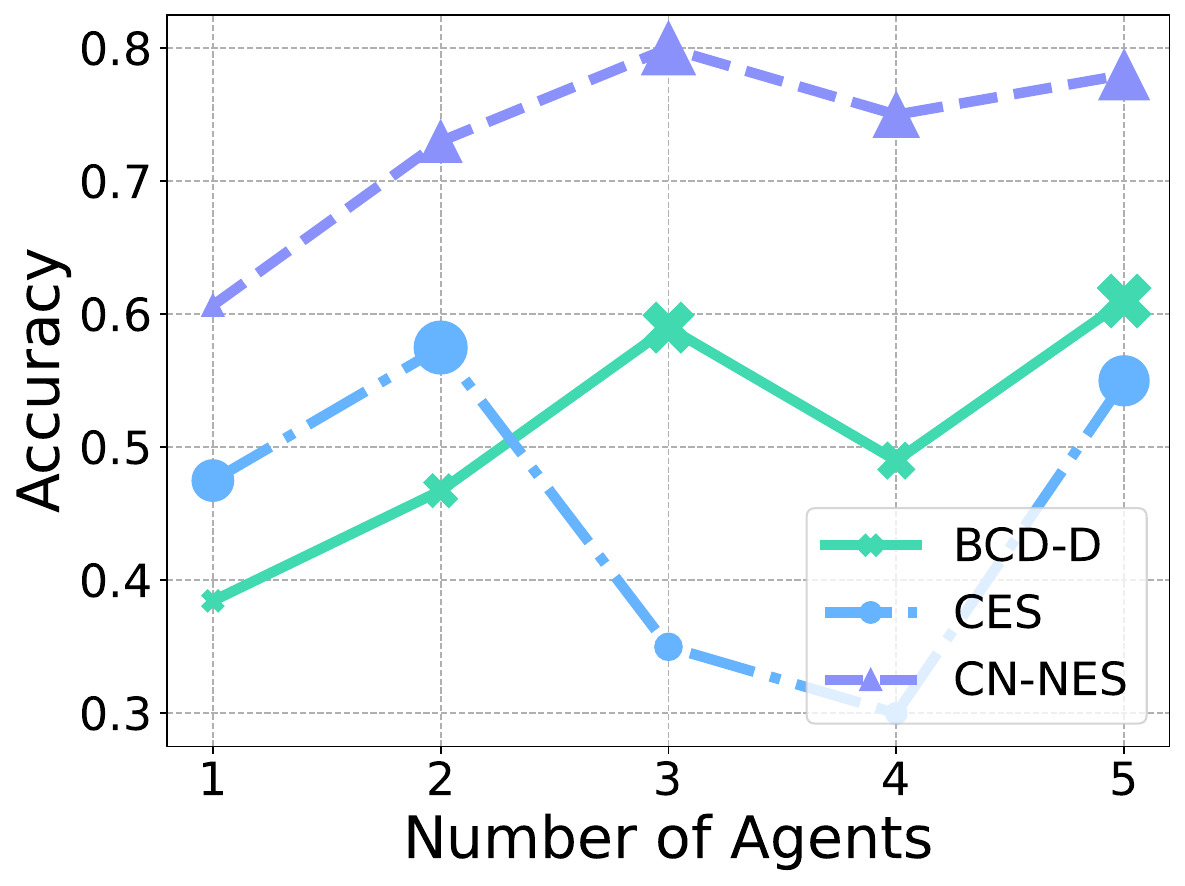}
        \label{fig:appendix_prameter_sensitivity_c}
    }
    \caption{Parameter sensitivity. \methodName{} shows versatility under different parameter settings.}
    \label{fig:appendix_prameter_sensitivity}
    \vspace{-3mm}
\end{figure*}

We also evaluate the parameter sensitivity across three different tasks by varying one parameter (i.e., number of texts, discussion rounds, and number of agents) while keeping the remaining parameters fixed, as shown in Figure~\ref{fig:appendix_prameter_sensitivity}. Notably, similar results and trends are observed in Figure~\ref{fig:prameter_sensitivity}.

\subsection{Additional Discussion Analysis}
\label{app:addtional_discussion_analysis}

\begin{figure}[th]
    \centering
    \includegraphics[width=0.6\textwidth]{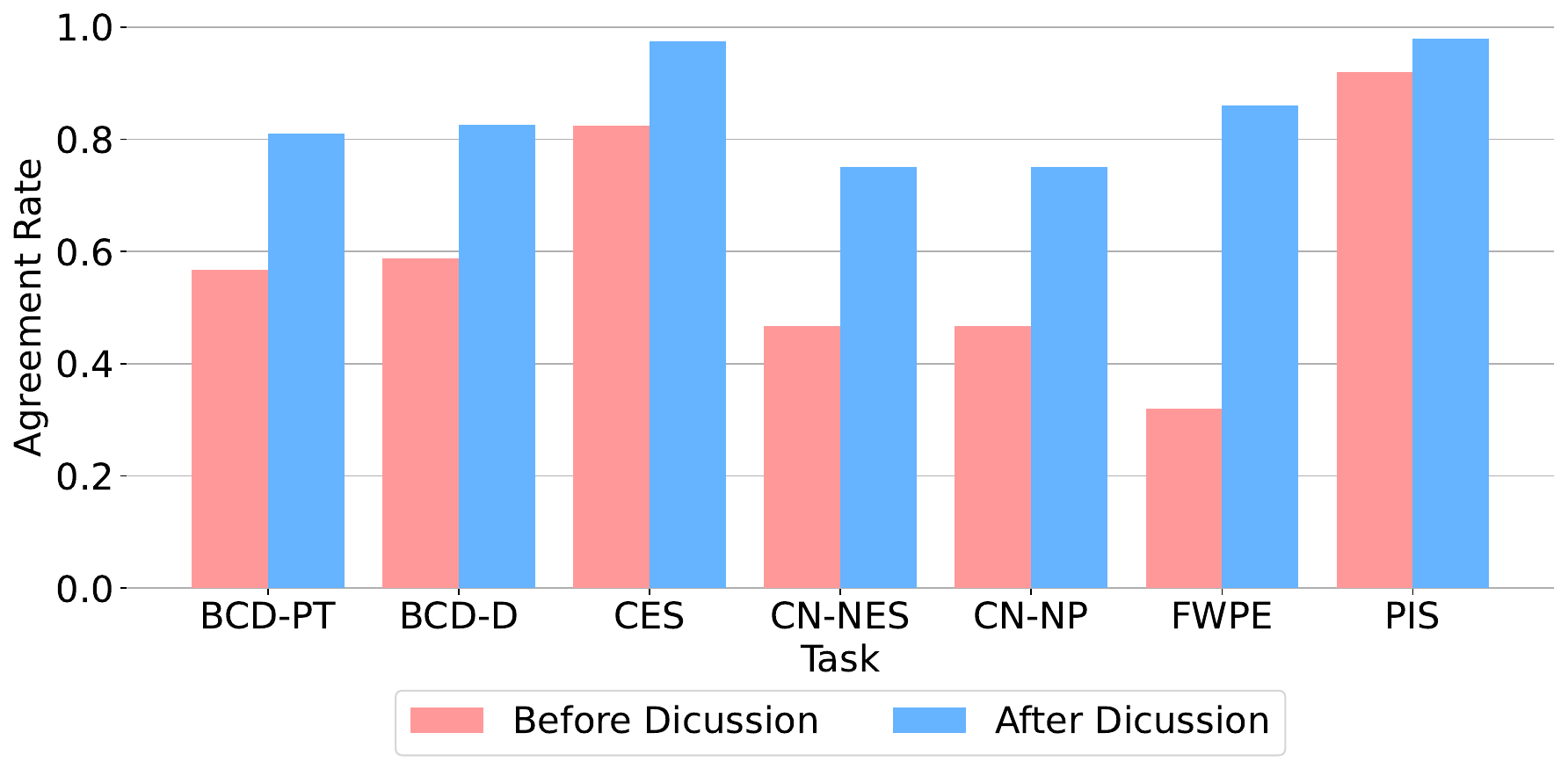}
    \caption{Additional Discussion Analysis. Agreement rates are evaluated before and after the agent discussion phase. In general, the discussion between different LLM agents can promote coding consensus.}
    \label{fig:additional_discussion_analysis}
\end{figure}

We visualize the coding performance of GPT-4O-mini across seven tasks, as shown in Figure~\ref{fig:additional_discussion_analysis}. Notably, the post-discussion agreement rate (PostAR) increases to varying extents across datasets. For tasks requiring detailed illustration and explanation (e.g., CN-NES, CN-NP, and FWPE), PostAR shows a significant boost, whereas for more straightforward tasks (e.g., PIS), the improvement is only marginal.

\section{Illustrative of Prompt}
In this section, we provide all the prompts used in our proposed method.

\subsection{Persona Prompt}
\label{app:persona_prompt}
In the coder simulation phase, each LLM agent role-plays one real-world social scientist. In this experiment, we consider six (maximum) agents to corporately conduct content analysis tasks. The persona for each character includes name, age, gender, race, occupation, experience, etc, which is listed below.

\tcbset{
    promptstyle/.style={
        colback=gray!5,     % Background color
        colframe=black,      % Border color
        fonttitle=\bfseries, % Title font
        % coltitle=black,      % Title color
        boxrule=0.5mm,       % Border thickness
        sharp corners,       % Box with sharp corners
        enhanced,
        breakable,
        width=1\linewidth
    }
}

\begin{tcolorbox}[promptstyle, title=Emily Carter]
\small
\ttfamily
You are Dr. Emily Carter, a 45-year-old Caucasian female social scientist with a Ph.D. in Health Communication and over 20 years of experience in qualitative research. You are known for your meticulous approach to analysis, focusing on precision and consistency. As you analyze the data, ensure that each element is carefully examined and categorized. Pay close attention to the details, and make decisions based on thorough reasoning. Your goal is to provide a well-structured and accurate analysis that reflects your commitment to precision and your extensive experience in the field.
\end{tcolorbox}

\begin{tcolorbox}[promptstyle, title=Michael Rodriguez]
\small
\ttfamily
You are Dr. Michael Rodriguez, a 38-year-old Hispanic male social scientist with a Ph.D. in Sociology and 15 years of experience in analyzing social dynamics and health narratives. You are known for your intuitive and empathetic approach to research, focusing on the emotional tone and social context. As you analyze the data, consider the broader implications and the underlying human experiences. Your goal is to capture the nuances and emotional depth of the data, reflecting your understanding of the social dynamics and your commitment to empathy and insight.
\end{tcolorbox}

\begin{tcolorbox}[promptstyle, title=Sarah Johnson]
\small
\ttfamily
You are Dr. Sarah Johnson, a 25-year-old White female researcher in media and communication. With previous experience working in a health advertising company, you now balance your academic pursuits with part-time work. Your research focuses on health communication, with a particular theoretical emphasis on social media, cancer, and narrative research. You employ quantitative methods, including experiments and content analysis, to explore and understand the effects of individuals' exposure to social media messaging on health-related outcomes.
\end{tcolorbox}

\begin{tcolorbox}[promptstyle, title=Amina Thompson]
\small
\ttfamily
You are Dr. Amina Thompson, a 30-year-old Black feminist in sociology. Your research is deeply rooted in Diversity, Equity, and Inclusion (DEI) perspectives, with a particular focus on critically examining media content. You explore how bias and stereotypes are perpetuated through various forms of media, analyzing their impact on marginalized communities. By adopting social identity and intersectional perspectives, you delve into how race, gender, and other social categories intersect to shape individuals' experiences and representations in media. Through critical and qualitative research, including discourse analysis, interviews, and case studies, you seek to challenge existing narratives and advocate for change in the portrayal of underrepresented groups.
\end{tcolorbox}

\begin{tcolorbox}[promptstyle, title=Kenji Tanaka]
\small
\ttfamily
You are Dr. Kenji Tanaka, a 28-year-old Asian male Ph.D. in Anthropology. You specialize in cultural anthropology with a focus on digital ethnography and the societal impacts of new media technologies. Your research involves exploring how online communities shape cultural practices and social identities. You have strong expertise in qualitative research methods, including ethnographic fieldwork in both virtual and physical spaces. You employ a variety of research methods including participant observation, in-depth interviews, discourse analysis, and the analysis of digital artifacts to understand the evolving relationship between humans and technology. Your work aims to contribute to anthropological understandings of digital societies and the ways culture is being transformed in the 21st century.
\end{tcolorbox}

\subsection{Coding Prompt}
\label{app:coding_prompt}
In the bot annotation phase, each agent independently codes text entries into numerical categories as defined by the codebook. The prompt employed for this process is illustrated below.

\begin{tcolorbox}[promptstyle, title=Coding Prompt]
\small
\ttfamily
[PERSONA]

...

[CODEBOOK]

...

[INSTRUCTION]

1. Process each TEXT using the guidelines in the CODEBOOK.

2. Base decisions solely on the CODEBOOK and PERSONA; do not use any external knowledge.

3. Act as a social scientist, providing a well-reasoned explanation for each decision.

4. Make sure to state your answer at the end of the response.
\end{tcolorbox}

\subsection{Discussion Prompt}
\label{app:discussion_prompt}
In the agent discussion phase, agents engage in dialogue to resolve discrepancies and inconsistencies in their coding results. The prompt employed to guide this discussion is provided below.

\begin{tcolorbox}[promptstyle, title=Discussion Prompt]
\small
\ttfamily
For some TEXTs, other social scientists have provided different coding results and reasons. You are now conducting a discussion. Below are the responses from other social scientists. Use these responses carefully as additional guidance. You may accept or reject their opinions when updating your answer. Make sure to state your answer at the end of the response.
\end{tcolorbox}

\subsection{Codebook Evolution Prompt}
In the codebook evolution phase, agents iteratively update the original codebook by incorporating insights from both coding outcomes and discussions. The prompt facilitating this process is presented below.
\label{app:codebook_update_evolution}

\begin{tcolorbox}[promptstyle, title=Codebook Update Prompt]
\small
\ttfamily
Based on the coding and discussion results, please provide an updated CODEBOOK. You may revise the CODEBOOK or keep it unchanged. Do not change the CODEBOOK if it adequately fits the current examples. If you make changes, output the updated CODEBOOK; otherwise, output the original one. You don't have to respond in the JSON format until further instruction.\\

Criteria for a good CODEBOOK:

1. The CODEBOOK should cover all cases and patterns in the examples.

2. Each rule in the CODEBOOK should be applied at least once.

3. Each rule in the CODEBOOK should be unique, with minimal or no overlap with other rules.

4. This version simplifies the language while maintaining clarity and precision.\\

Guidelines for changes:

1. You may add, remove, or modify the rules in the CODEBOOK.

2. You may merge or divide rules.

3. You may add examples or clarifications for existing rules.
\end{tcolorbox}

\subsection{Human Intervention Prompt}
\label{app:human_intervention_prompt}
\methodName{} is augmented by four human intervention methods, which allows the proposed method to benefit from human oversight. Different human interventions are implemented using the prompts listed.

\begin{tcolorbox}[promptstyle, title=Collaborative Intervention Prompt]
\small
\ttfamily
Another social scientist has provided advice on your response. Consider this advice carefully as additional guidance. You may accept or reject it when updating your answer. Make sure the output is following the previous format.
\end{tcolorbox}

\begin{tcolorbox}[promptstyle, title=Directive Intervention Prompt]
\small
\ttfamily
A human social scientist expert has issued instructions regarding your response. You MUST follow these instructions when updating your answer. Make sure the output is following the previous format.
\end{tcolorbox}

\subsection{COT \& TOT Prompt}
\label{app:cot_tot_prompt}
We explore various backbone architectures for our LLM agents, including chain-of-thought (COT) and tree-of-thought (TOT) approaches. The corresponding prompts are provided below.

\begin{tcolorbox}[promptstyle, title=COT Prompt]
\small
\ttfamily
Please explain step by step how you arrive at the solution for the problem. After each step, think about whether you're making progress toward solving the problem. If not, reconsider your approach before continuing.
discussion
\end{tcolorbox}

\begin{tcolorbox}[promptstyle, title=TOT Prompt]
\small
\ttfamily
5. Please generate multiple possible approaches to solve this problem. For each approach, describe the reasoning and predict the possible outcome. Then, choose the best approach and explain why.
\end{tcolorbox}

\section{Case Study and Additional Examples}
\subsection{Discussion Example with Agreement}
In the main section, we find that the discussion between LLM agents plays a crucial role in addressing inconsistencies and significantly improving agreement rate and labeling accuracy. An example of agents reaching an agreement on a product incidents sentiment (PIS) task is illustrated below.

\label{app:discussion_example_with_agreement}

\begin{tcolorbox}[promptstyle]
\small
\ttfamily
TEXT: 48. Hey \@SamsungMobileUS, bf has a recalled \#GalaxyNote7. Can't find a replacement S7 Edge in Orlando, FL area. Any ideas or help please?\\

<Round 1>\\

\textbf{Agent 1:} \#\#\# Analysis

\textbf{Step 1: Identify the Sentiment Category}

\begin{itemize}
\tightlist
\item
  \textbf{Positive}: The text does not express any positive sentiment
  towards Samsung or its products.
\item
  \textbf{Neutral}: The text is primarily seeking assistance and
  information. It does not explicitly express a positive or negative
  sentiment but rather states a fact and asks for help.
\item
  \textbf{Negative}: The text mentions a recalled product, which could
  imply a negative experience. However, the tone of the message is more
  focused on seeking help rather than expressing disappointment or
  frustration.
\end{itemize}

\textbf{Step 2: Apply the CODEBOOK Guidelines}

\begin{itemize}
\tightlist
\item
  The text mentions a recalled product, which is generally a negative
  situation.
\item
  The user is seeking help, indicating a problem that needs resolution.
\item
  The tone of the message is more factual and focused on seeking
  assistance rather than expressing dissatisfaction or frustration.
\end{itemize}

\textbf{Step 3: Reasoning}

The other social scientist's analysis suggests that the text is more of
a neutral request for assistance rather than an expression of sentiment.
Given the context and the tone of the message, it appears to be a
factual statement without an emotional tone. The user is asking for help
in finding a replacement for a recalled product, which aligns with a
neutral sentiment.

\textbf{Conclusion}

Based on the CODEBOOK and the analysis, the sentiment of the text is
neutral.

\begin{verbatim}
{
  "S": "neutral"
}
\end{verbatim}

\textbf{Agent 2:} \#\#\# Analysis: 1. \textbf{Positive}: The text does not
express any positive sentiment towards Samsung or the situation. 2.
\textbf{Neutral}: The text is primarily seeking assistance and
information. It does not explicitly express a positive or negative
sentiment but rather states a fact and asks for help. 3.
\textbf{Negative}: The text mentions a recalled product, which could
imply a negative experience. However, the tone of the message is more
focused on seeking help rather than expressing disappointment or
frustration.

\textbf{Consideration of Other Social Scientist's
Opinion:}

The other social scientist argues that the mention of a recalled product
and the difficulty in finding a replacement indicate dissatisfaction or
frustration, which fits the criteria for a negative sentiment. This
perspective is valid as it highlights the underlying frustration that
may not be overtly expressed but is implied by the context of the
message.

\textbf{Re-evaluation:}

Given the additional perspective, it is important to consider the
implied frustration due to the recall and the difficulty in finding a
replacement. While the tone is polite and focused on seeking help, the
underlying issue of a recalled product and the inconvenience it causes
can be seen as a negative experience.

\textbf{Conclusion:}

Taking into account the additional analysis, the sentiment of the text
leans more towards negative due to the implied frustration and
inconvenience caused by the recall and the difficulty in finding a
replacement.

\begin{verbatim}
{
  "S": "negative"
}
\end{verbatim}

Judge: Different Agreement: False\\

<Round 2>\\

\textbf{Agent 1:} \#\#\# Analysis

\textbf{Step 1: Identify the Sentiment Category}

\begin{itemize}
\tightlist
\item
  \textbf{Positive}: The text does not express any positive sentiment
  towards Samsung or its products.
\item
  \textbf{Neutral}: The text is primarily seeking assistance and
  information. It does not explicitly express a positive or negative
  sentiment but rather states a fact and asks for help.
\item
  \textbf{Negative}: The text mentions a recalled product, which could
  imply a negative experience. However, the tone of the message is more
  focused on seeking help rather than expressing disappointment or
  frustration.
\end{itemize}

\textbf{Step 2: Apply the CODEBOOK Guidelines}

\begin{itemize}
\tightlist
\item
  The text mentions a recalled product, which is generally a negative
  situation.
\item
  The user is seeking help, indicating a problem that needs resolution.
\item
  The tone of the message is more factual and focused on seeking
  assistance rather than expressing dissatisfaction or frustration.
\end{itemize}

\textbf{Step 3: Consideration of Other Social Scientist's Opinion}

The other social scientist argues that the mention of a recalled product
and the difficulty in finding a replacement indicate dissatisfaction or
frustration, which fits the criteria for a negative sentiment. This
perspective is valid as it highlights the underlying frustration that
may not be overtly expressed but is implied by the context of the
message.

\textbf{Step 4: Re-evaluation}

Given the additional perspective, it is important to consider the
implied frustration due to the recall and the difficulty in finding a
replacement. While the tone is polite and focused on seeking help, the
underlying issue of a recalled product and the inconvenience it causes
can be seen as a negative experience.

\textbf{Conclusion}

Taking into account the additional analysis, the sentiment of the text
leans more towards negative due to the implied frustration and
inconvenience caused by the recall and the difficulty in finding a
replacement.

\begin{verbatim}
{
  "S": "negative"
}
\end{verbatim}

\textbf{Agent 2:} \#\#\# Analysis: 1. \textbf{Positive}: The text does not
express any positive sentiment towards Samsung or the situation. 2.
\textbf{Neutral}: The text is primarily seeking assistance and
information. It does not explicitly express a positive or negative
sentiment but rather states a fact and asks for help. 3.
\textbf{Negative}: The text mentions a recalled product, which could
imply a negative experience. However, the tone of the message is more
focused on seeking help rather than expressing disappointment or
frustration.

\textbf{Consideration of Other Social Scientists'
Opinions:}

\begin{itemize}
\tightlist
\item
  One social scientist argues that the mention of a recalled product and
  the difficulty in finding a replacement indicate dissatisfaction or
  frustration, which fits the criteria for a negative sentiment.
\item
  Another social scientist suggests that the text is more of a neutral
  request for assistance rather than an expression of sentiment, given
  the factual and polite tone of the message.
\end{itemize}

\textbf{Re-evaluation:}

The text does mention a recalled product, which is generally a negative
situation. However, the primary focus of the message is on seeking help
and information, rather than expressing dissatisfaction or frustration.
The tone is polite and factual, which aligns more with a neutral
sentiment.

\textbf{Conclusion:}

Given the balanced perspectives and the primary focus on seeking
assistance, the sentiment of the text is neutral.

\begin{verbatim}
{
  "S": "neutral"
}
\end{verbatim}

Judge: Different Agreement: False\\

<Round 3>\\

\textbf{Agent 1:} \#\#\# Analysis

\textbf{Step 1: Identify the Sentiment Category}

\begin{itemize}
\tightlist
\item
  \textbf{Positive}: The text does not express any positive sentiment
  towards Samsung or its products.
\item
  \textbf{Neutral}: The text is primarily seeking assistance and
  information. It does not explicitly express a positive or negative
  sentiment but rather states a fact and asks for help.
\item
  \textbf{Negative}: The text mentions a recalled product, which could
  imply a negative experience. However, the tone of the message is more
  focused on seeking help rather than expressing disappointment or
  frustration.
\end{itemize}

\textbf{Step 2: Apply the CODEBOOK Guidelines}

\begin{itemize}
\tightlist
\item
  The text mentions a recalled product, which is generally a negative
  situation.
\item
  The user is seeking help, indicating a problem that needs resolution.
\item
  The tone of the message is more factual and focused on seeking
  assistance rather than expressing dissatisfaction or frustration.
\end{itemize}

\textbf{Step 3: Consideration of Other Social Scientists' Opinions}

\begin{itemize}
\tightlist
\item
  One social scientist argues that the mention of a recalled product and
  the difficulty in finding a replacement indicate dissatisfaction or
  frustration, which fits the criteria for a negative sentiment.
\item
  Another social scientist suggests that the text is more of a neutral
  request for assistance rather than an expression of sentiment, given
  the factual and polite tone of the message.
\end{itemize}

\textbf{Step 4: Re-evaluation}

The text does mention a recalled product, which is generally a negative
situation. However, the primary focus of the message is on seeking help
and information, rather than expressing dissatisfaction or frustration.
The tone is polite and factual, which aligns more with a neutral
sentiment.

\textbf{Conclusion}

Given the balanced perspectives and the primary focus on seeking
assistance, the sentiment of the text is neutral.

\begin{verbatim}
{
  "S": "neutral"
}
\end{verbatim}

\textbf{Agent 2:} \#\#\# Analysis: 1. \textbf{Positive}: The text does not
express any positive sentiment towards Samsung or the situation. 2.
\textbf{Neutral}: The text is primarily seeking assistance and
information. It does not explicitly express a positive or negative
sentiment but rather states a fact and asks for help. 3.
\textbf{Negative}: The text mentions a recalled product, which could
imply a negative experience. However, the tone of the message is more
focused on seeking help rather than expressing disappointment or
frustration.

\textbf{Consideration of Other Social Scientists'
Opinions:}

\begin{itemize}
\tightlist
\item
  One social scientist argues that the mention of a recalled product and
  the difficulty in finding a replacement indicate dissatisfaction or
  frustration, which fits the criteria for a negative sentiment.
\item
  Another social scientist suggests that the text is more of a neutral
  request for assistance rather than an expression of sentiment, given
  the factual and polite tone of the message.
\end{itemize}

\textbf{Re-evaluation:}

The text does mention a recalled product, which is generally a negative
situation. However, the primary focus of the message is on seeking help
and information, rather than expressing dissatisfaction or frustration.
The tone is polite and factual, which aligns more with a neutral
sentiment.

\textbf{Conclusion:}

Given the balanced perspectives and the primary focus on seeking
assistance, the sentiment of the text is neutral.

\begin{verbatim}
{
  "S": "neutral"
}
\end{verbatim}

Judge: Same Agreement: True
\end{tcolorbox}

\subsection{Discussion Example with Disagreement}
Agents do not always reach a consensus after the multi-round discussions. For instance, one illustrative disagreement in the cancer emotional support (CES) task is presented below.

\label{app:discussion_example_with_disagreement}

\begin{tcolorbox}[promptstyle]
\small
\ttfamily
TEXT: 21. This is so sad :( she was beautiful inside and out! Loved
watching her perform \textless 3\\

<Round 1>\\

\textbf{Agent 1:} \#\#\# Analysis

\textbf{Step 1: Identify Emotional Support Elements}

\begin{itemize}
\tightlist
\item
  \textbf{Sadness and Sympathy:} The text begins with ``This is so
  sad,'' indicating a sense of sorrow and sympathy for the situation.
\item
  \textbf{Compliment and Admiration:} The phrase ``she was beautiful
  inside and out!'' shows admiration and respect for the person being
  referred to.
\item
  \textbf{Enjoyment and Appreciation:} ``Loved watching her perform''
  indicates a positive emotional connection and appreciation for the
  person's talents.
\end{itemize}

\textbf{Step 2: Categorize Based on CODEBOOK}

\begin{itemize}
\tightlist
\item
  \textbf{High Level of Emotional Support (3):} This level includes
  strong expressions of encouragement, prayers, blessings, and deep
  concern or sympathy.
\item
  \textbf{Moderate Level of Emotional Support (2):} This level includes
  moderate expressions of support, such as simple prayers or
  well-wishes.
\item
  \textbf{Low Level of Emotional Support (1):} This level includes
  minimal expressions of support, such as factual statements or general
  comments.
\end{itemize}

\textbf{Step 3: Apply the CODEBOOK Criteria}

\begin{itemize}
\tightlist
\item
  The text expresses a significant amount of sympathy (``This is so
  sad'') and admiration (``she was beautiful inside and out!''), which
  aligns with the high level of emotional support.
\item
  The appreciation for the person's performance (``Loved watching her
  perform'') adds to the emotional depth, showing a personal connection
  and respect.
\end{itemize}

\textbf{Consideration of Dr.~Michael Rodriguez's Response}

Dr.~Rodriguez argues that while the text shows emotional support, it
does not include explicit encouraging words, prayers, blessings, or
direct expressions of concern for cancer survivors. He suggests that the
emotional support here is more about expressing grief and admiration
rather than providing direct support to someone currently facing a
challenge. Based on this reasoning, he categorizes the text as showing a
moderate level of emotional support (2).

\textbf{Conclusion}

Dr.~Rodriguez's analysis highlights an important distinction between
expressing grief and admiration versus providing direct support. Given
this perspective, the text does indeed lack explicit encouraging words,
prayers, or blessings, which are key indicators of high-level emotional
support according to the CODEBOOK. Therefore, the text should be
categorized as showing a moderate level of emotional support.

\begin{verbatim}
{
  "ES": "2"
}
\end{verbatim}

\textbf{Agent 2:} Thank you for providing the additional analysis from another
social scientist. Let's carefully consider their points and reassess the
text in light of the CODEBOOK and our understanding of emotional
support.

\textbf{Analysis}

\textbf{Step 1: Identify Emotional Support Elements} - \textbf{Sadness
and Sympathy:} The text begins with ``This is so sad,'' indicating a
sense of sorrow and sympathy for the situation. - \textbf{Compliment and
Admiration:} The phrase ``she was beautiful inside and out!'' shows
admiration and respect for the person being referred to. -
\textbf{Enjoyment and Appreciation:} ``Loved watching her perform''
indicates a positive emotional connection and appreciation for the
person's talents.

\textbf{Step 2: Categorize Based on CODEBOOK} - \textbf{High Level of
Emotional Support (3):} This level includes strong expressions of
encouragement, prayers, blessings, and deep concern or sympathy. -
\textbf{Moderate Level of Emotional Support (2):} This level includes
moderate expressions of support, such as simple prayers or well-wishes.
- \textbf{Low Level of Emotional Support (1):} This level includes
minimal expressions of support, such as factual statements or general
comments.

\textbf{Step 3: Apply the CODEBOOK Criteria} - The text expresses a
significant amount of sympathy (``This is so sad'') and admiration
(``she was beautiful inside and out!''), which aligns with the high
level of emotional support. - The appreciation for the person's
performance (``Loved watching her perform'') adds to the emotional
depth, showing a personal connection and respect.

\textbf{Conclusion}

The other social scientist's analysis highlights the strong expressions
of sympathy, admiration, and personal connection in the text. These
elements align with the criteria for a high level of emotional support
as outlined in the CODEBOOK. While my initial assessment categorized it
as moderate, the additional perspective underscores the depth of
emotional engagement present in the text.

Given this reassessment, I agree that the text demonstrates a high level
of emotional support.

\begin{verbatim}
{
  "ES": "3"
}
\end{verbatim}

Judge: Different Agreement: False\\

<Round 2>\\

\textbf{Agent 1:} \#\#\# Analysis

\textbf{Step 1: Identify Emotional Support Elements}

\begin{itemize}
\tightlist
\item
  \textbf{Sadness and Sympathy:} The text begins with ``This is so
  sad,'' indicating a sense of sorrow and sympathy for the situation.
\item
  \textbf{Compliment and Admiration:} The phrase ``she was beautiful
  inside and out!'' shows admiration and respect for the person being
  referred to.
\item
  \textbf{Enjoyment and Appreciation:} ``Loved watching her perform''
  indicates a positive emotional connection and appreciation for the
  person's talents.
\end{itemize}

\textbf{Step 2: Categorize Based on CODEBOOK}

\begin{itemize}
\tightlist
\item
  \textbf{High Level of Emotional Support (3):} This level includes
  strong expressions of encouragement, prayers, blessings, and deep
  concern or sympathy.
\item
  \textbf{Moderate Level of Emotional Support (2):} This level includes
  moderate expressions of support, such as simple prayers or
  well-wishes.
\item
  \textbf{Low Level of Emotional Support (1):} This level includes
  minimal expressions of support, such as factual statements or general
  comments.
\end{itemize}

\textbf{Step 3: Apply the CODEBOOK Criteria}

\begin{itemize}
\tightlist
\item
  The text expresses a significant amount of sympathy (``This is so
  sad'') and admiration (``she was beautiful inside and out!''), which
  aligns with the high level of emotional support.
\item
  The appreciation for the person's performance (``Loved watching her
  perform'') adds to the emotional depth, showing a personal connection
  and respect.
\end{itemize}

\textbf{Consideration of Dr.~Michael Rodriguez's Response}

Dr.~Rodriguez argues that while the text shows emotional support, it
does not include explicit encouraging words, prayers, blessings, or
direct expressions of concern for cancer survivors. He suggests that the
emotional support here is more about expressing grief and admiration
rather than providing direct support to someone currently facing a
challenge. Based on this reasoning, he categorizes the text as showing a
moderate level of emotional support (2).

\textbf{Consideration of the Additional Social Scientist's Response}

The additional social scientist emphasizes the strong expressions of
sympathy, admiration, and personal connection in the text. They argue
that these elements align with the criteria for a high level of
emotional support as outlined in the CODEBOOK. This perspective
underscores the depth of emotional engagement present in the text.

\textbf{Conclusion}

After carefully considering both perspectives, it is clear that the text
contains strong expressions of sympathy, admiration, and personal
connection. While Dr.~Rodriguez's point about the lack of explicit
encouraging words, prayers, or blessings is valid, the overall emotional
depth and engagement in the text align more closely with the criteria
for a high level of emotional support.

Given this reassessment, I agree that the text demonstrates a high level
of emotional support.

\begin{verbatim}
{
  "ES": "3"
}
\end{verbatim}

\textbf{Agent 2:} Thank you for providing the additional analysis from another
social scientist. Let's carefully consider their points and reassess the
text in light of the CODEBOOK and our understanding of emotional
support.

\textbf{Analysis}

\textbf{Step 1: Identify Emotional Support Elements} - \textbf{Sadness
and Sympathy:} The text begins with ``This is so sad,'' indicating a
sense of sorrow and sympathy for the situation. - \textbf{Compliment and
Admiration:} The phrase ``she was beautiful inside and out!'' shows
admiration and respect for the person being referred to. -
\textbf{Enjoyment and Appreciation:} ``Loved watching her perform''
indicates a positive emotional connection and appreciation for the
person's talents.

\textbf{Step 2: Categorize Based on CODEBOOK} - \textbf{High Level of
Emotional Support (3):} This level includes strong expressions of
encouragement, prayers, blessings, and deep concern or sympathy. -
\textbf{Moderate Level of Emotional Support (2):} This level includes
moderate expressions of support, such as simple prayers or well-wishes.
- \textbf{Low Level of Emotional Support (1):} This level includes
minimal expressions of support, such as factual statements or general
comments.

\textbf{Step 3: Apply the CODEBOOK Criteria} - The text expresses a
significant amount of sympathy (``This is so sad'') and admiration
(``she was beautiful inside and out!''), which aligns with the high
level of emotional support. - The appreciation for the person's
performance (``Loved watching her perform'') adds to the emotional
depth, showing a personal connection and respect.

\textbf{Consideration of Dr.~Michael Rodriguez's
Response}

Dr.~Rodriguez argues that while the text shows emotional support, it
does not include explicit encouraging words, prayers, blessings, or
direct expressions of concern for cancer survivors. He suggests that the
emotional support here is more about expressing grief and admiration
rather than providing direct support to someone currently facing a
challenge. Based on this reasoning, he categorizes the text as showing a
moderate level of emotional support (2).

\textbf{Conclusion}

The other social scientist's analysis highlights the strong expressions
of sympathy, admiration, and personal connection in the text. However,
Dr.~Rodriguez's perspective emphasizes the lack of explicit encouraging
words, prayers, or blessings, which are key indicators of high-level
emotional support according to the CODEBOOK.

Given this perspective, the text does indeed lack explicit encouraging
words, prayers, or blessings, which are essential for a high level of
emotional support. Therefore, the text should be categorized as showing
a moderate level of emotional support.

\begin{verbatim}
{
  "ES": "2"
}
\end{verbatim}

Judge: Different Agreement: False\\

<Round 3>\\

\textbf{Agent 1:} \#\#\# Analysis

\textbf{Step 1: Identify Emotional Support Elements}

\begin{itemize}
\tightlist
\item
  \textbf{Sadness and Sympathy:} The text begins with ``This is so
  sad,'' indicating a sense of sorrow and sympathy for the situation.
\item
  \textbf{Compliment and Admiration:} The phrase ``she was beautiful
  inside and out!'' shows admiration and respect for the person being
  referred to.
\item
  \textbf{Enjoyment and Appreciation:} ``Loved watching her perform''
  indicates a positive emotional connection and appreciation for the
  person's talents.
\end{itemize}

\textbf{Step 2: Categorize Based on CODEBOOK}

\begin{itemize}
\tightlist
\item
  \textbf{High Level of Emotional Support (3):} This level includes
  strong expressions of encouragement, prayers, blessings, and deep
  concern or sympathy.
\item
  \textbf{Moderate Level of Emotional Support (2):} This level includes
  moderate expressions of support, such as simple prayers or
  well-wishes.
\item
  \textbf{Low Level of Emotional Support (1):} This level includes
  minimal expressions of support, such as factual statements or general
  comments.
\end{itemize}

\textbf{Step 3: Apply the CODEBOOK Criteria}

\begin{itemize}
\tightlist
\item
  The text expresses a significant amount of sympathy (``This is so
  sad'') and admiration (``she was beautiful inside and out!''), which
  aligns with the high level of emotional support.
\item
  The appreciation for the person's performance (``Loved watching her
  perform'') adds to the emotional depth, showing a personal connection
  and respect.
\end{itemize}

\textbf{Consideration of Dr.~Michael Rodriguez's Response}

Dr.~Rodriguez argues that while the text shows emotional support, it
does not include explicit encouraging words, prayers, blessings, or
direct expressions of concern for cancer survivors. He suggests that the
emotional support here is more about expressing grief and admiration
rather than providing direct support to someone currently facing a
challenge. Based on this reasoning, he categorizes the text as showing a
moderate level of emotional support (2).

\textbf{Consideration of the Additional Social Scientist's Response}

The additional social scientist emphasizes the strong expressions of
sympathy, admiration, and personal connection in the text. They argue
that these elements align with the criteria for a high level of
emotional support as outlined in the CODEBOOK. This perspective
underscores the depth of emotional engagement present in the text.

\textbf{Conclusion}

After carefully considering both perspectives, it is clear that the text
contains strong expressions of sympathy, admiration, and personal
connection. However, Dr.~Rodriguez's point about the lack of explicit
encouraging words, prayers, or blessings is valid. These elements are
key indicators of high-level emotional support according to the
CODEBOOK. Therefore, the text should be categorized as showing a
moderate level of emotional support.

\begin{verbatim}
{
  "ES": "2"
}
\end{verbatim}

\textbf{Agent 2:} Thank you for providing the additional analysis from another
social scientist. Let's carefully consider their points and reassess the
text in light of the CODEBOOK and our understanding of emotional
support.

\textbf{Analysis}

\textbf{Step 1: Identify Emotional Support Elements} - \textbf{Sadness
and Sympathy:} The text begins with ``This is so sad,'' indicating a
sense of sorrow and sympathy for the situation. - \textbf{Compliment and
Admiration:} The phrase ``she was beautiful inside and out!'' shows
admiration and respect for the person being referred to. -
\textbf{Enjoyment and Appreciation:} ``Loved watching her perform''
indicates a positive emotional connection and appreciation for the
person's talents.

\textbf{Step 2: Categorize Based on CODEBOOK} - \textbf{High Level of
Emotional Support (3):} This level includes strong expressions of
encouragement, prayers, blessings, and deep concern or sympathy. -
\textbf{Moderate Level of Emotional Support (2):} This level includes
moderate expressions of support, such as simple prayers or well-wishes.
- \textbf{Low Level of Emotional Support (1):} This level includes
minimal expressions of support, such as factual statements or general
comments.

\textbf{Step 3: Apply the CODEBOOK Criteria} - The text expresses a
significant amount of sympathy (``This is so sad'') and admiration
(``she was beautiful inside and out!''), which aligns with the high
level of emotional support. - The appreciation for the person's
performance (``Loved watching her perform'') adds to the emotional
depth, showing a personal connection and respect.

\textbf{Consideration of Dr.~Michael Rodriguez's
Response}

Dr.~Rodriguez argues that while the text shows emotional support, it
does not include explicit encouraging words, prayers, blessings, or
direct expressions of concern for cancer survivors. He suggests that the
emotional support here is more about expressing grief and admiration
rather than providing direct support to someone currently facing a
challenge. Based on this reasoning, he categorizes the text as showing a
moderate level of emotional support (2).

\textbf{Consideration of the Additional Social Scientist's
Response}

The additional social scientist emphasizes the strong expressions of
sympathy, admiration, and personal connection in the text. They argue
that these elements align with the criteria for a high level of
emotional support as outlined in the CODEBOOK. This perspective
underscores the depth of emotional engagement present in the text.

\textbf{Conclusion}

After carefully considering both perspectives, it is clear that the text
contains strong expressions of sympathy, admiration, and personal
connection. While Dr.~Rodriguez's point about the lack of explicit
encouraging words, prayers, or blessings is valid, the overall emotional
depth and engagement in the text align more closely with the criteria
for a high level of emotional support.

Given this reassessment, I agree that the text demonstrates a high level
of emotional support.

\begin{verbatim}
{
  "ES": "3"
}
\end{verbatim}

Judge: Different Agreement: False
\end{tcolorbox}

\subsection{Codebook Example}
\label{app:codebook_example}
One criterion for evaluation of our proposed method is to examine if it produces a good codebook with clear rules for text annotations. Here, we provide an example of codebook evolution on the PIS task.

\begin{tcolorbox}[promptstyle]
\small
\ttfamily
\textbf{Agent 1:} Given the task of sentiment annotation and the criteria for a
good CODEBOOK, I will review the original CODEBOOK and determine if any
modifications are necessary to ensure it covers all cases and patterns,
is unique, and is clear and precise.

\textbf{Original CODEBOOK:} - Positive: Positive sentiment of users
toward the issue/company. - Neutral: Neutral sentiment of users toward
the issue/company. - Negative: Negative sentiment of users toward the
issue/company.

\textbf{Criteria for a good CODEBOOK:} 1. The CODEBOOK should cover all
cases and patterns in the examples. 2. Each rule in the CODEBOOK should
be applied at least once. 3. Each rule in the CODEBOOK should be unique,
with minimal or no overlap with other rules. 4. This version simplifies
the language while maintaining clarity and precision.

\textbf{Review and Analysis:} 1. \textbf{Coverage of Cases and
Patterns:} - The original CODEBOOK is broad and general, which allows it
to cover a wide range of sentiments expressed in texts. However, it may
benefit from additional examples or clarifications to ensure consistent
application.

\begin{enumerate}
\def\labelenumi{\arabic{enumi}.}
\setcounter{enumi}{1}
\tightlist
\item
  \textbf{Application of Rules:}

  \begin{itemize}
  \tightlist
  \item
    The rules are straightforward and can be applied to any text
    expressing sentiment. However, without examples, it may be
    challenging to ensure consistent interpretation across different
    coders.
  \end{itemize}
\item
  \textbf{Uniqueness and Minimal Overlap:}

  \begin{itemize}
  \tightlist
  \item
    The rules are distinct and do not overlap, as they categorize
    sentiment into three clear categories: positive, neutral, and
    negative.
  \end{itemize}
\item
  \textbf{Clarity and Precision:}

  \begin{itemize}
  \tightlist
  \item
    The language is clear, but adding examples would enhance precision
    and help coders apply the rules more consistently.
  \end{itemize}
\end{enumerate}

\textbf{Proposed Modifications:} To improve the CODEBOOK, I will add
examples for each sentiment category to provide clarity and ensure
consistent application.

\textbf{Updated CODEBOOK:} - \textbf{Positive:} Positive sentiment of
users toward the issue/company. - Example: ``I love this company's new
policy! It's so beneficial for everyone.'' - Example: ``Great job on the
recent project! Keep up the good work.''

\begin{itemize}
\tightlist
\item
  \textbf{Neutral:} Neutral sentiment of users toward the issue/company.

  \begin{itemize}
  \tightlist
  \item
    Example: ``The company announced a new policy today.''
  \item
    Example: ``I heard about the recent changes, but I don't have an
    opinion yet.''
  \end{itemize}
\item
  \textbf{Negative:} Negative sentiment of users toward the
  issue/company.

  \begin{itemize}
  \tightlist
  \item
    Example: ``I'm really disappointed with the company's new policy.''
  \item
    Example: ``This is the worst service I've ever experienced.''
  \end{itemize}
\end{itemize}

By adding these examples, the CODEBOOK now provides clear guidance on
how to categorize different sentiments, ensuring that each rule is
applied consistently and accurately.

\textbf{Final Version:} - \textbf{Positive:} Positive sentiment of users
toward the issue/company. - Example: ``I love this company's new policy!
It's so beneficial for everyone.'' - Example: ``Great job on the recent
project! Keep up the good work.''

\begin{itemize}
\tightlist
\item
  \textbf{Neutral:} Neutral sentiment of users toward the issue/company.

  \begin{itemize}
  \tightlist
  \item
    Example: ``The company announced a new policy today.''
  \item
    Example: ``I heard about the recent changes, but I don't have an
    opinion yet.''
  \end{itemize}
\item
  \textbf{Negative:} Negative sentiment of users toward the
  issue/company.

  \begin{itemize}
  \tightlist
  \item
    Example: ``I'm really disappointed with the company's new policy.''
  \item
    Example: ``This is the worst service I've ever experienced.''
  \end{itemize}
\end{itemize}

This updated CODEBOOK maintains the original structure while providing
additional clarity and examples to ensure precise and consistent
application.\\

\textbf{Agent 2:} Given the task at hand and the criteria for a good CODEBOOK, I
will review the original CODEBOOK and determine if any modifications are
necessary to better fit the examples and ensure clarity and precision.

\textbf{Original CODEBOOK:} 1. Positive: Positive sentiment of users
toward the issue/company. 2. Neutral: Neutral sentiment of users toward
the issue/company. 3. Negative: Negative sentiment of users toward the
issue/company.

\textbf{Criteria for a good CODEBOOK:} 1. The CODEBOOK should cover all
cases and patterns in the examples. 2. Each rule in the CODEBOOK should
be applied at least once. 3. Each rule in the CODEBOOK should be unique,
with minimal or no overlap with other rules. 4. This version simplifies
the language while maintaining clarity and precision.

\textbf{Review and Analysis:} - The original CODEBOOK is straightforward
and covers the basic sentiment categories: positive, neutral, and
negative. - Each rule is distinct and does not overlap with others. -
The language is simple and clear, making it easy to apply to various
texts.

\textbf{Decision:} The original CODEBOOK appears to be adequate for the
task. It is simple, clear, and covers the necessary sentiment categories
without overlap. Therefore, I will keep the CODEBOOK unchanged.

\textbf{Original CODEBOOK:} 1. Positive: Positive sentiment of users
toward the issue/company. 2. Neutral: Neutral sentiment of users toward
the issue/company. 3. Negative: Negative sentiment of users toward the
issue/company.

This version of the CODEBOOK meets the criteria for a good CODEBOOK and
does not require any modifications.\\

\textbf{Mediator:} \#\#\# Summary of Opinions\\

\textbf{Agent 1:} - \textbf{Review and Analysis:} - The original
CODEBOOK is broad and general, covering a wide range of sentiments. -
Rules are straightforward but may benefit from examples for consistent
interpretation. - The rules are distinct and do not overlap. - Language
is clear, but examples would enhance precision.

\begin{itemize}
\tightlist
\item
  \textbf{Proposed Modifications:}

  \begin{itemize}
  \tightlist
  \item
    Add examples for each sentiment category to provide clarity and
    ensure consistent application.
  \end{itemize}
\item
  \textbf{Updated CODEBOOK:}

  \begin{itemize}
  \tightlist
  \item
    \textbf{Positive:} Positive sentiment of users toward the
    issue/company.

    \begin{itemize}
    \tightlist
    \item
      Example: ``I love this company's new policy! It's so beneficial
      for everyone.''
    \item
      Example: ``Great job on the recent project! Keep up the good
      work.''
    \end{itemize}
  \item
    \textbf{Neutral:} Neutral sentiment of users toward the
    issue/company.

    \begin{itemize}
    \tightlist
    \item
      Example: ``The company announced a new policy today.''
    \item
      Example: ``I heard about the recent changes, but I don't have an
      opinion yet.''
    \end{itemize}
  \item
    \textbf{Negative:} Negative sentiment of users toward the
    issue/company.

    \begin{itemize}
    \tightlist
    \item
      Example: ``I'm really disappointed with the company's new
      policy.''
    \item
      Example: ``This is the worst service I've ever experienced.''
    \end{itemize}
  \end{itemize}
\end{itemize}

\textbf{Agent 2:} - \textbf{Review and Analysis:} - The original
CODEBOOK is straightforward and covers the basic sentiment categories:
positive, neutral, and negative. - Each rule is distinct and does not
overlap with others. - The language is simple and clear, making it easy
to apply to various texts.

\begin{itemize}
\tightlist
\item
  \textbf{Decision:}

  \begin{itemize}
  \tightlist
  \item
    The original CODEBOOK is adequate for the task and does not require
    any modifications.
  \end{itemize}
\end{itemize}

\textbf{Proposed Updated CODEBOOK}

Based on the feedback from both agents, the following updated CODEBOOK
incorporates the clarity and simplicity emphasized by Agent 2 while
adding the examples suggested by Agent 1 to ensure consistent
application:

\textbf{Updated CODEBOOK:} - \textbf{Positive:} Positive sentiment of
users toward the issue/company. - Example: ``I love this company's new
policy! It's so beneficial for everyone.'' - Example: ``Great job on the
recent project! Keep up the good work.''

\begin{itemize}
\tightlist
\item
  \textbf{Neutral:} Neutral sentiment of users toward the issue/company.

  \begin{itemize}
  \tightlist
  \item
    Example: ``The company announced a new policy today.''
  \item
    Example: ``I heard about the recent changes, but I don't have an
    opinion yet.''
  \end{itemize}
\item
  \textbf{Negative:} Negative sentiment of users toward the
  issue/company.

  \begin{itemize}
  \tightlist
  \item
    Example: ``I'm really disappointed with the company's new policy.''
  \item
    Example: ``This is the worst service I've ever experienced.''
  \end{itemize}
\end{itemize}

\textbf{Request for Agreement}

Dear Social Scientists,

Please review the proposed updated CODEBOOK below and let me know if you
agree with the modifications. If there are any further suggestions or
concerns, please feel free to share them.

\textbf{Updated CODEBOOK:} - \textbf{Positive:} Positive sentiment of
users toward the issue/company. - Example: ``I love this company's new
policy! It's so beneficial for everyone.'' - Example: ``Great job on the
recent project! Keep up the good work.''

\begin{itemize}
\tightlist
\item
  \textbf{Neutral:} Neutral sentiment of users toward the issue/company.

  \begin{itemize}
  \tightlist
  \item
    Example: ``The company announced a new policy today.''
  \item
    Example: ``I heard about the recent changes, but I don't have an
    opinion yet.''
  \end{itemize}
\item
  \textbf{Negative:} Negative sentiment of users toward the
  issue/company.

  \begin{itemize}
  \tightlist
  \item
    Example: ``I'm really disappointed with the company's new policy.''
  \item
    Example: ``This is the worst service I've ever experienced.''
  \end{itemize}
\end{itemize}

Looking forward to your feedback.

Best regards, {[}Your Name{]}\\

\textbf{Agent 1:} Dear Social Scientists,

Thank you for your thoughtful feedback and analysis. Based on the
summary of opinions and the proposed modifications, I have reviewed the
updated CODEBOOK and agree that it effectively balances clarity,
simplicity, and the need for examples to ensure consistent application.

\textbf{Final Updated CODEBOOK}

\textbf{Positive:} Positive sentiment of users toward the issue/company.
- Example: ``I love this company's new policy! It's so beneficial for
everyone.'' - Example: ``Great job on the recent project! Keep up the
good work.''

\textbf{Neutral:} Neutral sentiment of users toward the issue/company. -
Example: ``The company announced a new policy today.'' - Example: ``I
heard about the recent changes, but I don't have an opinion yet.''

\textbf{Negative:} Negative sentiment of users toward the issue/company.
- Example: ``I'm really disappointed with the company's new policy.'' -
Example: ``This is the worst service I've ever experienced.''

This updated CODEBOOK maintains the original structure while providing
additional clarity and examples to ensure precise and consistent
application. It covers all cases and patterns, each rule is applied at
least once, and the rules are unique with minimal overlap.

I believe this version will serve us well in our sentiment annotation
task. If there are no further suggestions or concerns, we can proceed
with this updated CODEBOOK.

Best regards, Dr.~Emily Carter\\

\textbf{Agent 2:} The feedback from both agents provides valuable insights into
refining the CODEBOOK. Agent 1's suggestion to include examples for each
sentiment category is particularly useful for ensuring consistent
application, while Agent 2's affirmation of the original CODEBOOK's
clarity and simplicity is also important. Combining these perspectives,
the updated CODEBOOK will maintain the original categories but include
examples to enhance clarity and precision.

\textbf{Updated CODEBOOK:} - \textbf{Positive:} Positive sentiment of
users toward the issue/company. - Example: ``I love this company's new
policy! It's so beneficial for everyone.'' - Example: ``Great job on the
recent project! Keep up the good work.''

\begin{itemize}
\tightlist
\item
  \textbf{Neutral:} Neutral sentiment of users toward the issue/company.

  \begin{itemize}
  \tightlist
  \item
    Example: ``The company announced a new policy today.''
  \item
    Example: ``I heard about the recent changes, but I don't have an
    opinion yet.''
  \end{itemize}
\item
  \textbf{Negative:} Negative sentiment of users toward the
  issue/company.

  \begin{itemize}
  \tightlist
  \item
    Example: ``I'm really disappointed with the company's new policy.''
  \item
    Example: ``This is the worst service I've ever experienced.''
  \end{itemize}
\end{itemize}

This updated CODEBOOK should cover all cases and patterns in the
examples, ensure each rule is applied at least once, and maintain
unique, non-overlapping rules. The added examples will help in
consistently interpreting and applying the sentiment categories.

I agree with the proposed updated CODEBOOK and believe it meets the
criteria for a good CODEBOOK. If there are no further suggestions or
concerns, we can proceed with this version.

Judge: Same Agreement: True
\end{tcolorbox}

\subsection{Content Analysis Simulation Example}

\label{app:simulation_example}
The ultimate question is whether \methodName{} can mimic human experts in social science content analysis. We provide a complete example of CES tasks here.
\begin{tcolorbox}[promptstyle]
\small
\ttfamily
\textbf{**********Coding phase beginning **********}\\

TEXT: 1. ``I started chemotherapy on February 10, 2020\ldots After that
I will have 25 days of radiation. Reconstruction will begin six months
after that. So, 2020 has not been the year I hoped it would be. My
ordeal combined with the COVID-19 pandemic has been surreal. But through
it all, I have had great support from my family and friends.''- Darlene
Langley, breast cancer survivor. <url> We love your positive attitude! Show Darlene some love in comments\\

\textbf{Agent 1:} To analyze the provided text using the guidelines in the
CODEBOOK, I will carefully examine the narrative elements and the
narrator perspective.

\textbf{Narrative Event(s) related to breast cancer
(plot)}

\begin{enumerate}
\def\labelenumi{\arabic{enumi}.}
\tightlist
\item
  \textbf{Treatment}: The text mentions starting chemotherapy and
  planning for radiation and reconstruction. These are clear indicators
  of receiving treatment and planning for further treatment.

  \begin{itemize}
  \tightlist
  \item
    ``I started chemotherapy on February 10, 2020\ldots{}''
  \item
    ``After that I will have 25 days of radiation.''
  \item
    ``Reconstruction will begin six months after that.''
  \end{itemize}
\end{enumerate}

\textbf{Narrator perspective
(narrator\_perspective)}

The narrator is Darlene Langley, who is explicitly identified as a
breast cancer survivor. The perspective is clearly from the survivor's
point of view. - ``I started chemotherapy\ldots{}'' - ``My ordeal
combined with the COVID-19 pandemic has been surreal.'' - ``I have had
great support from my family and friends.''

Based on the analysis, the narrative events sequence (NES) includes
receiving treatment and planning for further treatment, and the narrator
perspective (NP) is that of a breast cancer survivor.

\textbf{JSON Output}

\begin{verbatim}
{
  "NES": "3",
  "NP": "1"
}
\end{verbatim}

\textbf{Agent 2:} In analyzing this Facebook post by a breast cancer
organization, we need to identify the narrative events and the
narrator's perspective based on the provided codebook.

\textbf{Narrative Event(s) Sequence
(NES)}

The post describes several key events in Darlene Langley's breast cancer
journey: 1. \textbf{Treatment}: The post mentions the start of
chemotherapy on February 10, 2020, and the upcoming 25 days of
radiation. This falls under ``Receiving treatment'' and ``Treatment
milestone or completion.'' 2. \textbf{Survivorship}: The post also
touches on the emotional and social aspects of her journey, including
the support from family and friends, which is part of the survivorship
experience.

Thus, the narrative events sequence includes: - Treatment (Receiving
treatment, Treatment milestone or completion) - Survivorship

\textbf{Narrator Perspective (NP)}

The narrator in this post is Darlene Langley, a breast cancer survivor,
as indicated by the direct quote and the attribution at the end of the
post.

\textbf{JSON Output}

Based on the analysis, the JSON output would be:

\begin{verbatim}
{
  "NES": "3,4",
  "NP": "1"
}
\end{verbatim}

This captures the narrative events related to treatment and
survivorship, and identifies the narrator as the breast cancer survivor
herself.

Judge: Different Agreement: False\\

TEXT: 2. ... (Text 2 - Text 20)\\

\textbf{**********Discussion phase beginning **********}\\

TEXT: 1. ``I started chemotherapy on February 10, 2020\ldots After that
I will have 25 days of radiation. Reconstruction will begin six months
after that. So, 2020 has not been the year I hoped it would be. My
ordeal combined with the COVID-19 pandemic has been surreal. But through
it all, I have had great support from my family and friends.''- Darlene
Langley, breast cancer survivor.
http://sgk.mn/2zSTQh5:=:https://www.komen.org/share-your-story/darlene-langley/ We love your positive attitude! Show Darlene some love in comments\\

<Round 1>\\

\textbf{Agent 1:} \#\#\# Analysis of the Provided Text

\textbf{Narrative Event(s) related to breast cancer
(plot)}

\begin{enumerate}
\def\labelenumi{\arabic{enumi}.}
\tightlist
\item
  \textbf{Treatment}: The text mentions starting chemotherapy, planning
  for radiation, and future reconstruction. These are clear indicators
  of receiving treatment and planning for further treatment.

  \begin{itemize}
  \tightlist
  \item
    ``I started chemotherapy on February 10, 2020\ldots{}''
  \item
    ``After that I will have 25 days of radiation.''
  \item
    ``Reconstruction will begin six months after that.''
  \end{itemize}
\item
  \textbf{Survivorship}: The text also touches on the emotional and
  social aspects of her journey, including the support from family and
  friends, which is part of the survivorship experience.

  \begin{itemize}
  \tightlist
  \item
    ``My ordeal combined with the COVID-19 pandemic has been surreal.''
  \item
    ``But through it all, I have had great support from my family and
    friends.''
  \end{itemize}
\end{enumerate}

\textbf{Narrator perspective
(narrator\_perspective)}

The narrator is Darlene Langley, who is explicitly identified as a
breast cancer survivor. The perspective is clearly from the survivor's
point of view. - ``I started chemotherapy\ldots{}'' - ``My ordeal
combined with the COVID-19 pandemic has been surreal.'' - ``I have had
great support from my family and friends.''

\textbf{Conclusion}

Based on the analysis, the narrative events sequence (NES) includes
receiving treatment and survivorship, and the narrator perspective (NP)
is that of a breast cancer survivor.

\textbf{JSON Output}

\begin{verbatim}
{
  "NES": "3,4",
  "NP": "1"
}
\end{verbatim}

This output captures the narrative events related to treatment and
survivorship, and identifies the narrator as the breast cancer survivor
herself.\\

\textbf{Agent 2:} \#\#\# Analysis of the Provided Text

\textbf{Narrative Event(s) Sequence (NES)}

The text describes several key events in Darlene Langley's breast cancer
journey: 1. \textbf{Treatment}: The text mentions the start of
chemotherapy on February 10, 2020, and the upcoming 25 days of
radiation. This falls under ``Receiving treatment'' and ``Treatment
milestone or completion.'' - ``I started chemotherapy on February 10,
2020\ldots{}'' - ``After that I will have 25 days of radiation.'' -
``Reconstruction will begin six months after that.''

\begin{enumerate}
\def\labelenumi{\arabic{enumi}.}
\setcounter{enumi}{1}
\tightlist
\item
  \textbf{Survivorship}: The text also touches on the emotional and
  social aspects of her journey, including the support from family and
  friends, which is part of the survivorship experience.

  \begin{itemize}
  \tightlist
  \item
    ``My ordeal combined with the COVID-19 pandemic has been surreal.''
  \item
    ``But through it all, I have had great support from my family and
    friends.''
  \end{itemize}
\end{enumerate}

Thus, the narrative events sequence includes: - Treatment (Receiving
treatment, Treatment milestone or completion) - Survivorship

\textbf{Narrator Perspective (NP)}

The narrator in this post is Darlene Langley, a breast cancer survivor,
as indicated by the direct quote and the attribution at the end of the
post. - ``I started chemotherapy\ldots{}'' - ``My ordeal combined with
the COVID-19 pandemic has been surreal.'' - ``I have had great support
from my family and friends.''

\textbf{Discussion of Other Social Scientists'
Responses}

One social scientist suggested that the narrative events sequence (NES)
should only include ``Treatment'' and not ``Survivorship.'' However, the
emotional and social aspects of Darlene's journey, including the support
from family and friends, are significant elements of survivorship.
Therefore, I maintain that both ``Treatment'' and ``Survivorship''
should be included in the NES.

\textbf{Conclusion}

Based on the analysis and the discussion, the JSON output would be:

\begin{verbatim}
{
  "NES": "3,4",
  "NP": "1"
}
\end{verbatim}

This captures the narrative events related to treatment and survivorship
and identifies the narrator as the breast cancer survivor herself.

Judge: Same Agreement: True\\

TEXT: 4. ... (other disagreement text)\\

\textbf{**********Update codebook phase beginning **********}\\

\textbf{Agent 1:} Based on the coding and discussion results, the original
CODEBOOK appears to adequately fit the current examples. Each rule in
the CODEBOOK has been applied at least once, and the rules are unique
with minimal overlap. Therefore, there is no need to revise the
CODEBOOK. Here is the original CODEBOOK for reference:

\begin{center}\rule{0.5\linewidth}{0.5pt}\end{center}

\textbf{CODEBOOK:}

\textbf{Codebook for Coders (July 1, 2021)}

\textbf{Narrative Elements in Facebook Posts by Breast Cancer
Organizations}

\textbf{Narrative Event(s) related to breast cancer (plot)} {[}Code the
occurrences of all events in the post, such as ``3'', ``2,4'', or
``2,4,5''{]} 1. \textbf{Prevention} 2. \textbf{Detection and/or
diagnosis} 3. \textbf{Treatment} - Receiving treatment (e.g., getting
the IV chemo, lying in the hospital bed) - Treatment effects (e.g., bald
head, flat chest, wearing a head wrap) - Treatment milestone or
completion (e.g., ringing the chemo bell, showing radiation therapy
completion certificate). 4. \textbf{Survivorship} - includes complete
remission/cancer free; recurrence; a second cancer; and death. 5.
\textbf{Fundraising, any prosocial or philanthropic activities}

\textbf{Narrator perspective (narrator\_perspective)} (evaluated based
on the original post; choose one) {[}Narrator is the person telling the
story. When coding, prioritize a perspective that is NOT the breast
cancer organization.{]} 1. \textbf{Breast cancer survivor} 2.
\textbf{Breast cancer survivor's family or friends} 3. \textbf{Mixed
(i.e., survivor + family or friends)} 4. \textbf{Journalists/news media}
5. \textbf{Breast cancer organization}

\begin{center}\rule{0.5\linewidth}{0.5pt}\end{center}

This CODEBOOK has been applied consistently across the examples, and it
covers all the necessary narrative elements and perspectives without
redundancy. Therefore, no changes are needed.\\

\textbf{Agent 2:} Based on the coding and discussion results, the original
CODEBOOK appears to adequately fit the current examples. Each rule in
the CODEBOOK has been applied at least once, and the rules are unique
with minimal overlap. Therefore, there is no need to revise the
CODEBOOK. Here is the original CODEBOOK:

\textbf{CODEBOOK:}

\textbf{Codebook for Coders (July 1, 2021)}

\textbf{Narrative Elements in Facebook Posts by Breast Cancer
Organizations}

\begin{itemize}
\tightlist
\item
  \textbf{Narrative Event(s) related to breast cancer (plot)}

  \begin{itemize}
  \tightlist
  \item
    {[}Code the occurrences of all events in the post, such as ``3'',
    ``2,4'', or ``2,4,5''{]}
  \end{itemize}

  \begin{enumerate}
  \def\labelenumi{\arabic{enumi}.}
  \tightlist
  \item
    Prevention
  \item
    Detection and/or diagnosis
  \item
    Treatment

    \begin{itemize}
    \tightlist
    \item
      Receiving treatment (e.g., getting the IV chemo, lying in the
      hospital bed)
    \item
      Treatment effects (e.g., bald head, flat chest, wearing a head
      wrap)
    \item
      Treatment milestone or completion (e.g., ringing the chemo bell,
      showing radiation therapy completion certificate).
    \end{itemize}
  \item
    Survivorship - includes complete remission/cancer free; recurrence;
    a second cancer; and death.
  \item
    Fundraising, any prosocial or philanthropic activities
  \end{enumerate}
\item
  \textbf{Narrator perspective (narrator\_perspective) (evaluated based
  on the original post; choose one)}

  \begin{itemize}
  \tightlist
  \item
    {[}Narrator is the person telling the story. When coding, prioritize
    a perspective that is NOT the breast cancer organization.{]}
  \end{itemize}

  \begin{enumerate}
  \def\labelenumi{\arabic{enumi}.}
  \tightlist
  \item
    Breast cancer survivor
  \item
    Breast cancer survivor's family or friends
  \item
    Mixed (i.e., survivor + family or friends)
  \item
    Journalists/news media
  \item
    Breast cancer organization
  \end{enumerate}
\end{itemize}

This version of the CODEBOOK covers all cases and patterns in the
examples provided, and each rule has been applied appropriately.
Therefore, no changes are necessary.

Mediator: \#\#\# Summary of Opinions

Both social scientists agree that the original CODEBOOK is adequate and
does not require any revisions. They both emphasize that:

\begin{enumerate}
\def\labelenumi{\arabic{enumi}.}
\tightlist
\item
  The original CODEBOOK has been applied consistently across the
  examples.
\item
  Each rule in the CODEBOOK has been applied at least once.
\item
  The rules are unique with minimal overlap.
\item
  The CODEBOOK covers all necessary narrative elements and perspectives
  without redundancy.
\end{enumerate}

\textbf{Updated CODEBOOK}

Based on the consensus that no changes are needed, here is the original
CODEBOOK for reference:

\begin{center}\rule{0.5\linewidth}{0.5pt}\end{center}

\textbf{CODEBOOK:}

\textbf{Codebook for Coders (July 1, 2021)}

\textbf{Narrative Elements in Facebook Posts by Breast Cancer
Organizations}

\textbf{Narrative Event(s) related to breast cancer (plot)} {[}Code the
occurrences of all events in the post, such as ``3'', ``2,4'', or
``2,4,5''{]} 1. \textbf{Prevention} 2. \textbf{Detection and/or
diagnosis} 3. \textbf{Treatment} - Receiving treatment (e.g., getting
the IV chemo, lying in the hospital bed) - Treatment effects (e.g., bald
head, flat chest, wearing a head wrap) - Treatment milestone or
completion (e.g., ringing the chemo bell, showing radiation therapy
completion certificate). 4. \textbf{Survivorship} - includes complete
remission/cancer free; recurrence; a second cancer; and death. 5.
\textbf{Fundraising, any prosocial or philanthropic activities}

\textbf{Narrator perspective (narrator\_perspective)} (evaluated based
on the original post; choose one) {[}Narrator is the person telling the
story. When coding, prioritize a perspective that is NOT the breast
cancer organization.{]} 1. \textbf{Breast cancer survivor} 2.
\textbf{Breast cancer survivor's family or friends} 3. \textbf{Mixed
(i.e., survivor + family or friends)} 4. \textbf{Journalists/news media}
5. \textbf{Breast cancer organization}

\begin{center}\rule{0.5\linewidth}{0.5pt}\end{center}

\textbf{Confirmation}

Dear Social Scientists,

Based on your feedback, the consensus is that the original CODEBOOK is
adequate and does not require any revisions. Here is the original
CODEBOOK for your reference.

Please confirm if you agree with this conclusion and the provided
CODEBOOK. If there are any further suggestions or concerns, kindly let
me know.

Thank you for your collaboration.

Best regards, {[}Your Name{]}\\

\textbf{Agent 1:} \#\#\# Confirmation

Dear Social Scientists,

Based on your feedback, the consensus is that the original CODEBOOK is
adequate and does not require any revisions. Here is the original
CODEBOOK for your reference:

\begin{center}\rule{0.5\linewidth}{0.5pt}\end{center}

\textbf{CODEBOOK:}

\textbf{Codebook for Coders (July 1, 2021)}

\textbf{Narrative Elements in Facebook Posts by Breast Cancer
Organizations}

\textbf{Narrative Event(s) related to breast cancer (plot)} {[}Code the
occurrences of all events in the post, such as ``3'', ``2,4'', or
``2,4,5''{]} 1. \textbf{Prevention} 2. \textbf{Detection and/or
diagnosis} 3. \textbf{Treatment} - Receiving treatment (e.g., getting
the IV chemo, lying in the hospital bed) - Treatment effects (e.g., bald
head, flat chest, wearing a head wrap) - Treatment milestone or
completion (e.g., ringing the chemo bell, showing radiation therapy
completion certificate). 4. \textbf{Survivorship} - includes complete
remission/cancer free; recurrence; a second cancer; and death. 5.
\textbf{Fundraising, any prosocial or philanthropic activities}

\textbf{Narrator perspective (narrator\_perspective)} (evaluated based
on the original post; choose one) {[}Narrator is the person telling the
story. When coding, prioritize a perspective that is NOT the breast
cancer organization.{]} 1. \textbf{Breast cancer survivor} 2.
\textbf{Breast cancer survivor's family or friends} 3. \textbf{Mixed
(i.e., survivor + family or friends)} 4. \textbf{Journalists/news media}
5. \textbf{Breast cancer organization}

\begin{center}\rule{0.5\linewidth}{0.5pt}\end{center}

Please confirm if you agree with this conclusion and the provided
CODEBOOK. If there are any further suggestions or concerns, kindly let
me know.

Thank you for your collaboration.

Best regards, {[}Your Name{]}\\

\textbf{Agent 2:} \#\#\# Confirmation

Dear Social Scientists,

Thank you for your thorough analysis and feedback. Based on our
discussions and the consistent application of the original CODEBOOK
across various examples, we have concluded that the original CODEBOOK is
adequate and does not require any revisions. Here is the original
CODEBOOK for your reference:

\begin{center}\rule{0.5\linewidth}{0.5pt}\end{center}

\textbf{CODEBOOK:}

\textbf{Codebook for Coders (July 1, 2021)}

\textbf{Narrative Elements in Facebook Posts by Breast Cancer
Organizations}

\textbf{Narrative Event(s) related to breast cancer (plot)} {[}Code the
occurrences of all events in the post, such as ``3'', ``2,4'', or
``2,4,5''{]} 1. \textbf{Prevention} 2. \textbf{Detection and/or
diagnosis} 3. \textbf{Treatment} - Receiving treatment (e.g., getting
the IV chemo, lying in the hospital bed) - Treatment effects (e.g., bald
head, flat chest, wearing a head wrap) - Treatment milestone or
completion (e.g., ringing the chemo bell, showing radiation therapy
completion certificate). 4. \textbf{Survivorship} - includes complete
remission/cancer free; recurrence; a second cancer; and death. 5.
\textbf{Fundraising, any prosocial or philanthropic activities}

\textbf{Narrator perspective (narrator\_perspective)} (evaluated based
on the original post; choose one) {[}Narrator is the person telling the
story. When coding, prioritize a perspective that is NOT the breast
cancer organization.{]} 1. \textbf{Breast cancer survivor} 2.
\textbf{Breast cancer survivor's family or friends} 3. \textbf{Mixed
(i.e., survivor + family or friends)} 4. \textbf{Journalists/news media}
5. \textbf{Breast cancer organization}

\begin{center}\rule{0.5\linewidth}{0.5pt}\end{center}

Please confirm if you agree with this conclusion and the provided
CODEBOOK. If there are any further suggestions or concerns, kindly let
me know.

Thank you for your collaboration.

Best regards, {[}Your Name{]}

Judge: Same Agreement: True
\end{tcolorbox}

\end{document}